\documentclass[journal,transmag]{IEEEtran}
\usepackage{cite}
\usepackage{amsmath,amssymb,amsfonts}
\usepackage{algorithmic}
\usepackage{graphicx}
\usepackage{textcomp}
\usepackage{xcolor}
\usepackage{fancyhdr}
\usepackage[hyphens]{url}

\newcommand{\ignore}[1]{}
\usepackage[pass]{geometry}
\usepackage[normalem]{ulem}
\usepackage{hyperref}
\usepackage{color}
\usepackage{soul}
\usepackage{tikz}
\usepackage{array}
\usepackage{fixltx2e}
\usepackage{stfloats}
\usepackage{times}
\usepackage{verbatim}
\usepackage{subfiles}
\usepackage{booktabs}
\usepackage{multirow}
\usepackage{caption}
\usepackage{subcaption}
\usepackage{floatrow}
\usepackage[flushleft]{threeparttable}
\usepackage[font=small,labelfont=bf]{caption}

\DeclareUnicodeCharacter{2215}{}
\ifCLASSINFOpdf
\else
\fi
\begin{document}
%
\title{A Survey of FPGA-Based Robotic Computing}


\author{\IEEEauthorblockN{Zishen Wan\textsuperscript{*}\IEEEauthorrefmark{1,2},
Bo Yu\textsuperscript{*}\IEEEauthorrefmark{3},
Thomas Yuang Li\IEEEauthorrefmark{3}, 
Jie Tang\IEEEauthorrefmark{4}, 
Yuhao Zhu\IEEEauthorrefmark{5},
Yu Wang\IEEEauthorrefmark{6},
Arijit Raychowdhury\IEEEauthorrefmark{1}, \\
and
Shaoshan Liu\IEEEauthorrefmark{3}}
\IEEEauthorblockA{\IEEEauthorrefmark{1}School of Electrical and Computer Engineering,
Georgia Institute of Technology, Atlanta, GA 30332 USA}
\IEEEauthorblockA{\IEEEauthorrefmark{2}John A. Paulson School of Engineering and Applied Sciences,
Harvard University, Cambridge, MA 02138 USA}
\IEEEauthorblockA{\IEEEauthorrefmark{3}PerceptIn Inc, Fremont, CA 94539 USA}
\IEEEauthorblockA{\IEEEauthorrefmark{4}School of Computer Science and Engineering, South China University of Technology, Guangzhou, Guangdong, China}
\IEEEauthorblockA{\IEEEauthorrefmark{5}Department of Computer Science, University of Rochester, Rochester, NY 14627 USA}
\IEEEauthorblockA{\IEEEauthorrefmark{6}Department of Electronic Engineering, Tsinghua University, Beijing, China}
\thanks{* These authors contributed equally to this work. 

Corresponding author: Shaoshan Liu (email: shaoshan.liu@perceptin.io).}}

%



\IEEEtitleabstractindextext{%
\begin{abstract}
Recent researches on robotics have shown significant improvement, spanning from algorithms, mechanics to hardware architectures. Robotics, including manipulators, legged robots, drones, and autonomous vehicles, are now widely applied in diverse scenarios. However, the high computation and data complexity of robotic algorithms pose great challenges to its applications. On the one hand, CPU platform is flexible to handle multiple robotic tasks. GPU platform has higher computational capacities and easy-to-use development frameworks, so they have been widely adopted in several applications. On the other hand, FPGA-based robotic accelerators are becoming increasingly competitive alternatives, especially in latency-critical and power-limited scenarios. With specialized designed hardware logic and algorithm kernels, FPGA-based accelerators can surpass CPU and GPU in performance and energy efficiency. In this paper, we give an overview of previous work on FPGA-based robotic accelerators covering different stages of the robotic system pipeline. An analysis of software and hardware optimization techniques and main technical issues is presented, along with some commercial and space applications, to serve as a guide for future work. 

\end{abstract}

\begin{IEEEkeywords}
Robotics, Autonomous Machines, Computer Architecture, FPGA, Space Exploration.
\end{IEEEkeywords}}

\maketitle

\IEEEdisplaynontitleabstractindextext

%
\IEEEpeerreviewmaketitle

\section{Introduction}
\label{sec:introduction}
Over the last decade, we have seen significant progress in the development of robotics, spanning from algorithms, mechanics to hardware platforms. Various robotic systems, like manipulators, legged robots, unmanned aerial vehicles, self-driving cars have been designed for search and rescue~\cite{qiantori2012emergency,ryan2005mode}, exploration~\cite{smolyanskiy2017toward,giusti2015machine}, package delivery~\cite{stolaroff2018energy}, entertainment~\cite{kim2018survey,jung2018direct} and more applications and scenarios. These robots are on the rise of demonstrating their full potential. Take drones, a type of aerial robots, as an example, the number of drones has grown by 2.83x between 2015 and 2019 based on the U.S. Federal Aviation Administration (FAA) report~\cite{FAA2020}. The registered number has reached 1.32 million in 2019, and the FFA expects this number will come to 1.59 billion by 2024. 

However, robotic systems are pretty complicated~\cite{liu2017creating,krishnan2020sky,krishnan2021machine}. They tightly integrate many technologies and algorithms, including sensing, perception, mapping, localization, decision making, control, etc. This complexity poses many challenges for the design of robotic edge computing systems \cite{liu2020autonomous}\cite{liu2019edge}. On the one hand, the robotic system needs to process an enormous amount of data in real-time. The incoming data often comes from multiple sensors and is highly heterogeneous. However, the robotic system usually has limited on-board resources, such as memory storage, bandwidth, and compute capabilities, making it hard to meet the real-time requirements. On the other hand, the current state-of-the-art robotic system usually has strict power constraints on the edge that cannot support the amount of computation required for performing tasks, such as 3D sensing, localization, navigation, and path planning. Therefore, the computation and storage complexity, as well as real-time and power constraints of the robotic system, hinder its wide application in latency-critical or power-limited scenarios \cite{liu2017computer}.

Therefore, it is essential to choose a proper compute platform for the robotic system. CPU and GPU are two widely used commercial compute platforms. CPU is designed to handle a wide range of tasks quickly and is often used to develop novel algorithms. A typical CPU can achieve 10-100 GFLOPS with below 1GOP/J power efficiency~\cite{guo2019dl}. In contrast, GPU is designed with thousands of processor cores running simultaneously, which enable massive parallelism. A typical GPU can perform up to 10 TOPS performance and become a good candidate for high-performance scenarios. Recently, benefiting in part from the better accessibility provided by CUDA/OpenCL, GPU has been predominantly used in many robotic applications. However,  conventional CPU and GPUs usually consume 10W to 100W of power, which are orders of magnitude higher than what is available on the resource-limited robotic system.

Besides CPU and GPU, FPGAs are attracting attention and becoming a platform candidate to achieve energy-efficient robotics tasks processing. FPGAs require little power and are often built into small systems with less memory. They have the ability to parallel computations massively and makes use of the properties of perception (e.g., stereo matching), localization (e.g., SLAM), and planning (e.g., graph search) kernels to remove additional logic and simplify the implementation. Taking into account hardware characteristics, several algorithms are proposed which can be run in a hardware-friendly way and achieve similar software performance. Therefore, FPGAs are possible to meet real-time requirements while achieving high energy efficiency compared to CPUs and GPUs.

Unlike the ASIC counterparts, FPGA technology provides the flexibility of on-site programming and re-programming without going through re-fabrication with a modified design. Partial Reconfiguration (PR) takes this flexibility one step further, allowing the modification of an operating FPGA design by loading a partial configuration file. Using PR, part of the FPGA can be reconfigured at runtime without compromising the integrity of the applications running on those parts of the device that are not being reconfigured. As a result, PR can allow different robotic applications to time-share part of an FPGA, leading to energy and performance efficiency, and making FPGA a suitable computing platform for dynamic and complex robotic workloads.

FPGAs have been successfully utilized in commercial autonomous vehicles. Particularly, over the past three years, PerceptIn has built and commercialized autonomous vehicles for micromobility, and PerceptIn's products have been deployed in China, US, Japan and Switzerland. In this paper, we review how PerceptIn developed its computing system by relying heavily on FPGAs, which perform not only heterogeneous sensor synchronizations, but also the acceleration of software components on the critical path. In addition, FPGAs are used heavily in space robotic applications, for FPGAs offered unprecedented flexibility and significantly reduced the design cycle and development cost. In this paper, we also delve into space-grade FPGAs for robotic applications.

The rest of paper is organized as follows: Section~\ref{sec:overview} introduces the basic workloads of the robotic system. Section~\ref{sec:perception}, \ref{sec:localization} and~\ref{sec:planning_control} reviews the various perception, localization and motion planning algorithms and their implementations on FPGA platforms. In section~\ref{sec:partial_reconfig}, we discuss about FPGA partial reconfiguration techniques. Section~\ref{sec:case_study1} and~\ref{sec:case_study2} present robotics FPGA applications in commercial and space areas. Section~\ref{sec:conclusion} concludes the paper.

\section{Overview of Robotics workloads}
\label{sec:overview}
\subsection{Overview}
\label{subsec:overview}
\begin{figure}[t!]
        \centering\includegraphics[width=0.75\columnwidth]{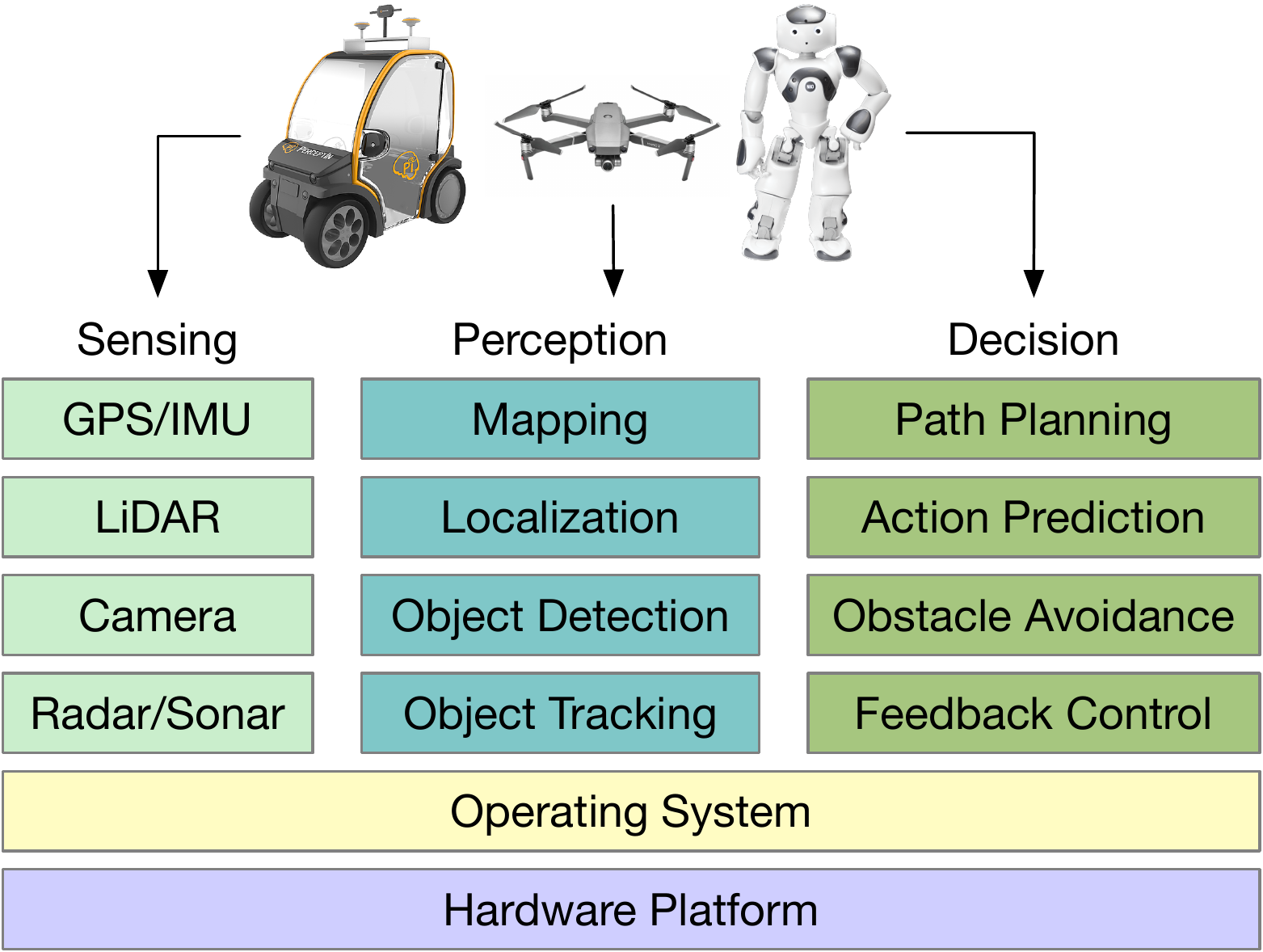}
        \caption{The stack of the robotic system.}
        \label{fig:robotics_system}
\end{figure}

Robotics is not one technology but rather an integration of many technologies. As shown in Fig~\ref{fig:robotics_system}, the stack of the robotic system consists of three major components: application workloads, including sensing, perception, localization, motion planning, and control; a software edge subsystem, including operating system and runtime layer; and computing hardware, including both microcontrollers and companion computers.

We focus on the robotic application workloads in this section. The application subsystem contains multiple algorithms that are used by the robot to extract meaningful information from raw sensor data to understand the environment and dynamically make decisions about its actions.

\subsection{Sensing}
\label{subsec:sensing}
The sensing stage is responsible for extracting meaningful information from the sensor raw data. To enable intelligent actions and improve reliability, the robot platform usually supports a wide range of sensors. The number and type of sensors are heavily dependent on the specifications of the workload and the capability of the onboard compute platform. The sensors can include the following:

\textbf{Cameras.} Cameras are usually used for object recognition and object tracking, such as lane detection in autonomous vehicles and obstacle detection in drones, etc. RGB-D camera can also be utilized to determine object distances and positions. Take autonomous vehicle as an example, the current system usually mounts eight or more 1080p cameras around the vehicle to detect, recognize and track objects in different directions, which can greatly improve safety. Usually, these cameras run at 60 Hz, which will process multiple gigabytes of raw data per second when combined.

\textbf{GNSS/IMU.} The global navigation satellite system (GNSS) and inertial measurement unit (IMU) system help the robot localize itself by reporting both inertial updates and an estimate of the global location at a high rate. Different robots have different requirements for localization sensing.  For instance, 10 Hz may be enough for low-speed mobile robots, but high-speed autonomous vehicles usually demand 30 Hz or higher for localization, and high-speed drones may need 100 Hz or more for localization, thus we are facing a broad spectrum of sensing speeds. Fortunately, different sensors have their own advantages and drawbacks. GNSS can enable fairly accurate localization, while it runs at only 10Hz, thus unable to provide real-time updates. By contrast, both accelerometer and gyroscope in IMU can run at 100-200 Hz, which can satisfy the real-time requirement. However, IMU suffers bias wandering over time or perturbation by some thermo-mechanical noise, which may lead to an accuracy degradation in the position estimates. By combining GNSS and IMU, we can get accurate and real-time updates for robots.

\textbf{LiDAR.} Light detection and ranging (LiDAR) is used for evaluating distance by illuminating the obstacles with laser light and measuring the reflection time. These pulses, along with other recorded data, can generate precise and three-dimensional information about the surrounding characteristics. LiDAR plays an important role in localization, obstacle detection and avoidance. As indicated in \cite{yu2020micro}, the choice of sensors dictates the algorithm and hardware design. Take autonomous driving as an instance, almost all the autonomous vehicle companies use LiDAR at the core of their technologies. Examples include Uber, Waymo, and Baidu. PerceptIn and Tesla are among the very few that do not use LiDAR and, instead, rely on cameras and vision-based systems, and in particular PerceptIn's data demonstrated that for the low-speed autonomous driving scenario, LiDAR processing is slower than camera-based vision processing, but increases the power consumption and cost.

\textbf{Radar and Sonar.} The Radio Detection and Ranging (Radar) and Sound Navigation and Ranging (Sonar) system is used to determine the distance and speed to a certain object, which usually serves as the last line of defense to avoid obstacles. Take autonomous vehicle as an example, a danger of collision may occur when near obstacles are detected, then the vehicle will apply brakes or turn to avoid obstacles. Compared to LiDAR, the Radar and Sonar system is cheaper and smaller, and their raw data is usually fed to the control processor directly without going through the main compute pipeline, which can be used to implement some urgent functions as swerving or applying the brakes.

\subsection{Perception}
The sensor data is then fed into the perception layer to sense the static and dynamic objects, and build a reliable and detailed representation of the robot’s environment using computer vision techniques (including deep learning).

The perception layer is responsible for object detection, segmentation and tracking. There are obstacles, lane dividers and other objects to detect. Traditionally, a detection pipeline starts with image pre-processing, followed by a region of interest detector and then a classifier that outputs detected objects. In 2005, Dalal and Triggs~\cite{1467360} proposed an algorithm based on histogram of orientation (HOG) and support vector machine (SVM) to model both the appearance and shape of the object under various condition. The goal of segmentation is to give the robot a structured understanding of its environment. Semantic segmentation is usually formulated as a graph labeling problem with vertices of the graph being pixels or super-pixels. Inference algorithms on graphical models such as conditional random field (CRF)~\cite{1315232,101007} are used. The goal of tracking is to estimate the trajectory of moving obstacles. Tracking can be formulated as a sequential Bayesian filtering problem by recursively running the prediction step and correction step. Tracking can also be formulated by tracking-by-detection handling with Markovian decision process (MDP)~\cite{7410891}, where an object detector is applied to consecutive frames and detected objects are linked across frames. 

In recent years, deep neural networks (DNN), also known as deep learning, have greatly affected computer vision and made significant progress in solving robot perception problems. Most state-of-the-art algorithms now apply one type of neural network based on convolution operation. Fast R-CNN~\cite{Girshick_2015}, Faster R-CNN~\cite{DBLP:journals/corr/RenHG015}, SSD~\cite{DBLP:journals/corr/LiuAESR15}, YOLO~\cite{DBLP:journals/corr/RedmonDGF15}, and YOLO9000~\cite{DBLP:journals/corr/RedmonF16} were used to get much better speed and accuracy in object detection. Most CNN-based semantic segmentation work is based on Fully Convolutional Networks (FCN)~\cite{DBLP:journals/corr/LongSD14}, and there are some recent work in spatial pyramid pooling network~\cite{DBLP:journals/corr/HeZR014} and pyramid scene parsing network (PSPNet)~\cite{DBLP:journals/corr/ZhaoSQWJ16} to combine global image-level information with the locally extracted feature. By using auxiliary natural images, a stacked autoencoder model can be trained offline to learn generic image features and then applied for online object tracking~\cite{DBLP:journals/corr/BertinettoVHVT16}.

\subsection{Localization}
The localization layer is responsible for aggregating data from various sensors to locate the robot in the environment model.

GNSS/IMU system is used for localization. The GNSS consist of several satellite systems, such as GPS, Galileo and BeiDou, which can provide accurate localization results but with a slow update rate. In comparison, IMU can provide a fast update with less accurate rotation and acceleration results. A mathematical filter, such as Kalman Filter, can be used to combine the advantages of the two and minimize the localization error and latency. However, this sole system has some problems, such as the signal may bounce off obstacles, introduce more noise, and fail to work in closed environments.

LiDAR and High-Definition (HD) maps are used for localization. LiDAR can generate point clouds and provide a shape description of the environment, while it is hard to differentiate individual points. HD map has a higher resolution compared to digital maps and makes the route familiar to the robot, where the key is to fuse different sensor information to minimize the errors in each grid cell. Once the HD map is built, a particle filter method can be applied to localize the robot in real-time correlated with LiDAR measurement. However, the LiDAR performance may be severely affected by weather conditions (e.g., rain, snow) and bring localization error.

Cameras are used for localization as well. The pipeline of vision-based localization is simplified as follows: 1) by triangulating stereo image pairs, a disparity map is obtained and used to derive depth information for each point; 2) by matching salient features between successive stereo image frames in order to establish correlations between feature points in different frames, the motion between the past two frames is estimated; and 3) by comparing the salient features against those in the known map, the current position of the robot is derived~\cite{1638022}. 

Apart from these techniques, sensor fusion strategy is also often utilized to combine multiple sensors for localization, which can improve the reliability and robustness of robot~\cite{montemerlo2008junior,6803933}.

\subsection{Planning and Control}
The planning and control layer is responsible for generating trajectory plans and passing the control commands based on the original and destination of the robot. Broadly, prediction and routing modules are also included here, where their outputs are fed into downstream planning and control layers as input. The prediction module is responsible for predicting the future behavior of surrounding objects identified by the perception layer. The routing module can be a lane-level routing based on lane segmentation of the HD maps for autonomous vehicles. 

Planning and Control layers usually include behavioral decision, motion planning and feedback control. The mission of the behavioral decision module is to make effective and safe decisions by leveraging all various input data sources. Bayesian models are becoming more and more popular and have been applied in recent works~\cite{katrakazas2015real,paden2016survey}. Among the Bayesian models, Markov Decision Process (MDP) and Partially Observable Markov Decision Process (POMDP) are the widely applied methods in modeling robot behavior. The task of motion planning is to generate a trajectory and send it to the feedback control for execution. The planned trajectory is usually specified and represented as a sequence of planned trajectory points, and each of these points contains attributes like location, time, speed, etc. Low-dimensional motion planning problems can be solved with grid-based algorithms (such as Dijkstra~\cite{deng2012fuzzy} or A*~\cite{hart1968formal}) or geometric algorithms. High-dimensional motion planning problems can be dealt with sampling-based algorithms, such as Rapidly-exploring Random Tree (RRT)~\cite{lavalle2001randomized} and Probabilistic Roadmap (PRM)~\cite{kavraki1996probabilistic}, which can avoid the problem of local minima. Reward-based algorithms, such as the Markov decision process (MDP), can also generate the optimal path by maximizing cumulative future rewards. The goal of feedback control is to track the difference between the actual pose and the pose on the predefined trajectory by continuous feedback. The most typical and widely used algorithm in robot feedback control is PID.

While optimization-based approaches enjoy mainstream appeal in solving motion planning and control problems, learning-based approaches~\cite{shalev2016long,gomez2012optimal,shalev2016safe,bojarski2016end,geng2017scenario} are becoming increasingly popular with recent developments in artificial intelligence. Learning-based methods, such as reinforcement learning, can naturally make full use of historical data and iteratively interact with the environment through actions to deal with complex scenarios. Some model the behavioral level decisions via reinforcement learning~\cite{shalev2016safe,geng2017scenario}, while other approaches directly work on motion planning trajectory output or even direct feedback control signals~\cite{gomez2012optimal}. Q-learning~\cite{watkins1992q}, Actor-Critic learning~\cite{konda2000actor}, policy gradient~\cite{kavraki1996probabilistic} are some popular algorithms in reinforcement learning.





\section{Perception on FPGA}
\label{sec:perception}
\subsection{Overview}
Perception is related to many robotic applications where sensory data and artificial intelligence techniques are involved. Examples of such applications include stereo matching, object detection, scene understanding, semantic classification, etc. The recent developments in machine learning, especially deep learning, have exposed robotic perception systems to more tasks. In this section, we will focus on the recent algorithms and FPGA implementations in the stereo vision system, which is one of the key components in the robotic perception stage.

Real-time and robust stereo vision systems are increasingly popular and widely used in many perception applications, e.g., robotics navigation, obstacle avoidance~\cite{hicks2013depth} and scene reconstruction~\cite{whelan2016elasticfusion,prisacariu2017infinitam,golodetz2018collaborative}. The purpose of stereo vision systems is to obtain 3D structure information of the scene using stereoscopic ranging techniques. The system usually has two cameras to capture images from two points of view within the same scenario. The disparities between the corresponding pixels in two stereo images are searched using stereo matching algorithms. Then the depth information can be calculated from the inverse of this disparity.

Throughout the whole pipeline, stereo matching is the bottleneck and time-consuming stage. The stereo matching algorithms can be mainly classified into two categories: local algorithms~\cite{perez2019fpga,yang2014depth,aguilar2016fpga,perez2016fpga,cocorullo2016efficient, santos2016scalable,ali2017exploring} and global algorithms~\cite{mccullagh2012real,li20173d,zha2016real,puglia2017real,kjaer2010two}. Local methods compute the disparities by only processing and matching the pixels around the points of interest within windows. They are fast and computationally-cheap, and the lack of pixel dependencies makes them suitable for parallel acceleration. However, they may suffer in textureless areas and occluded regions, which will result in incorrect disparities estimation.

In contrast, global methods compute the disparities by matching all other pixels and minimizing a global cost function. They can achieve much higher accuracy than local methods. However, they tend to come at high computation cost and require much more resources due to their large and irregular memory access as well as the sequential nature of algorithms, thus not suitable for real-time and low-power applications. Many research works in stereo systems focus on the speed and accuracy improvement of stereo matching algorithms, and some of the implementations are summarized in Tab.~\ref{tab:perception-table}

\subsection{Local Stereo Matching on FPGA}
Local algorithms are usually based on correlation, where the process involves finding matching pixels in the left and right image patches by aggregating costs within a specific region. There are many ways for cost aggregation, such as the sum of absolute differences (SAD)~\cite{wong2002sum}, the sum of squared differences (SSD)~\cite{hisham2015template}, normalized cross-correlation (NCC)~\cite{yoo2009fast}, and census transform (CT)~\cite{froba2004face}. Many FPGA implementations are based on these methods. Jin et al.~\cite{jin2009fpga} develop a real-time stereo vision system based on census rank transformation matching cost for 640×480 resolution images. Zhang et al.~\cite{zhang2011real} propose a real-time high definition stereo matching design on FPGA based on mini-census transform and cross-based cost aggregation, which achieves 60 fps at 1024×768 pixel stereo images. The implementation of Honegger et al.~\cite{honegger2012real} achieves 127 fps at 376×240 pixel resolution with 32 disparity levels based on block matching. Jin et al.~\cite{jin2014fast} further achieve 507.9 fps for 640×480 resolution images by applying fast local consistent dense stereo functions and cost aggregation. 

\subsection{Global Stereo Matching on FPGA}
\label{subsec:global_SM}
Global algorithms can provide state-of-the-art accuracy and disparity map quality, however, they are usually processed through high computational-intensive optimization techniques or massive convolutional neural networks, making them difficult to be deployed on resource-limited embedded systems for real-time applications. However, some works have attempted to implement global algorithms on FPGA for better performance. Park et al.~\cite{park2007real} present a trellis-based stereo matching system on FPGA with a low error rate and achieved 30 fps at 320×240 resolution with 128 disparity levels. Sabihuddin et al.~\cite{sabihuddin2008dynamic} implement a dynamic programming maximum likelihood (DPML) based hardware architecture for dense binocular disparity estimation and achieved 63.54 fps at 640×480 pixel resolution with 128 disparity levels. The design in Jin et al.~\cite{jin2012real} uses a tree-structured dynamic programming method, and achieves 58.7 fps at 640×480 resolution as well as a low error rate. Recently, some other adaptations of global approaches for FPGA-implementation have been proposed, such as cross-trees~\cite{zha2016real}, dynamic programming for DNA sequence alignment~\cite{puglia2017real}, and graph cuts~\cite{kamasaka2018fpga}, where all of these implementations achieve real-time processing.

\begin{table*}[t!]
\centering
\renewcommand*{\arraystretch}{1}
\resizebox{\columnwidth}{!}{%
\begin{tabular}{c|c|c|c|c|c|c|c|c}
\toprule[1.5pt]
\textbf{Algorithm} & \textbf{Reference} & \begin{tabular}[c]{@{}c@{}} \textbf{Frame Rate} \\  \textbf{(FPS)} \end{tabular}  & \begin{tabular}[c]{@{}c@{}} \textbf{Image Resolution} \\  \textbf{(Width $\times$ Height)} \end{tabular} & \begin{tabular}[c]{@{}c@{}} \textbf{Disparity} \\  \textbf{Level} \end{tabular} & 
\textbf{MDE/s} & \begin{tabular}[c]{@{}c@{}} \textbf{Power} \\  \textbf{(W)} \end{tabular}
& \begin{tabular}[c]{@{}c@{}} \textbf{Resource(\%)} \\  \textbf{Logic / BRAM} \end{tabular} &
\textbf{FPGA Platform} \\ \hline
\begin{tabular}[c]{@{}c@{}} Local \\  Stereo Matching \end{tabular} & 
\begin{tabular}[c]{@{}c@{}} Jin et al.~\cite{jin2009fpga} \\  Zhang et al.~\cite{zhang2011real} \\Honegger et al.~\cite{honegger2012real}\\ Jin et al.~\cite{jin2014fast} \end{tabular} &
\begin{tabular}[c]{@{}c@{}} 230 \\  60 \\127\\ 507.9 \end{tabular} &
\begin{tabular}[c]{@{}c@{}} 640 $\times$ 480 \\  1024 $\times$ 768 \\376 $\times$ 240\\ 640 $\times$ 480 \end{tabular} &
\begin{tabular}[c]{@{}c@{}} 64 \\  64 \\ 32 \\ 60 \end{tabular} &
\begin{tabular}[c]{@{}c@{}} 4522 \\  3020 \\ 367 \\ 9362 \end{tabular} &
\begin{tabular}[c]{@{}c@{}} -- \\  1.56 \\ 2.8 \\ 3.35 \end{tabular} &
\begin{tabular}[c]{@{}c@{}} 34.0 / 95.0 \\  61.8 / 67.0 \\ 49.0 / 68.0 \\ 81.0 / 39.7 \end{tabular} &
\begin{tabular}[c]{@{}c@{}} Xilinx Virtex-4 XC4VLX200-10 \\  Altera EP3SL150  \\ AItera Cyclone III EP3C80 \\ Xilinx Vertex-6 \end{tabular} 
 \\ \hline
\begin{tabular}[c]{@{}c@{}} Global \\  Stereo Matching \end{tabular} & 
\begin{tabular}[c]{@{}c@{}} Park et al.~\cite{park2007real} \\  Sabihuddin et al.~\cite{sabihuddin2008dynamic} \\ Jin et al.~\cite{jin2012real} \\ Zha et al.~\cite{zha2016real} \\ Puglia et al.~\cite{puglia2017real} \end{tabular} & 
\begin{tabular}[c]{@{}c@{}} 30 \\  63.54 \\ 32 \\ 30 \\ 30 \end{tabular} &
\begin{tabular}[c]{@{}c@{}} 320 $\times$ 240 \\  640 $\times$ 480 \\ 640 $\times$ 480 \\ 1920 $\times$ 1680 \\ 1024 $\times$ 768 \end{tabular} &
\begin{tabular}[c]{@{}c@{}} 128 \\  128 \\ 60 \\ 60 \\ 64 \end{tabular} &
\begin{tabular}[c]{@{}c@{}} 295 \\  2498 \\ 590 \\ 5806 \\ 1510 \end{tabular} &
\begin{tabular}[c]{@{}c@{}} -- \\  -- \\ 1.40 \\ -- \\ 0.17 \end{tabular} &
\begin{tabular}[c]{@{}c@{}} -- / -- \\  23.0 / 58.0 \\ 72.0 / 46.0 \\ 84.8 / 91.9 \\ 57.0 / 53.0 \end{tabular} &
\begin{tabular}[c]{@{}c@{}} Xilinx Virtex II pro-100 \\  Xilinx XC2VP100 \\ Xilinx XC4VLX160 \\ Xilinx Kintex 7 \\ Xilinx Virtex-7 XC7Z020CLG484-1 \\
\end{tabular} \\
\hline
\begin{tabular}[c]{@{}c@{}} Semi-Global \\  Stereo Matching \end{tabular} & 
\begin{tabular}[c]{@{}c@{}}  Banz et al.~\cite{banz2010real} \\ Wang et al.~\cite{wang2015real} \\ Cambuim et al.~\cite{cambuim2017hardware}  \\  Rahnama et al.~\cite{rahnama2018r3sgm} \\ Cambuim et al.~\cite{cambuim2019fpga} \\ Zhao et al.~\cite{zhao2020fp} \end{tabular} &
\begin{tabular}[c]{@{}c@{}} 37 \\  42.61 \\ 127  \\ 72 \\ 25 \\147 \end{tabular} &
\begin{tabular}[c]{@{}c@{}} 640 $\times$ 480 \\  1600 $\times$ 1200 \\ 1024 $\times$ 768  \\ 1242 $\times$ 375 \\ 1024 $\times$ 768 \\ 1242 $\times$ 375 \end{tabular} &
\begin{tabular}[c]{@{}c@{}} 128 \\ 128 \\ 128  \\ 128 \\ 256 \\ 64 \end{tabular} &
\begin{tabular}[c]{@{}c@{}} 1455 \\  10472 \\ 12784 \\ 4292 \\ 5033 \\ 4382 \end{tabular} &
\begin{tabular}[c]{@{}c@{}} 2.31 \\  2.79 \\ -- \\ 3.94 \\ 6.5 \\ 9.8 \end{tabular} &
\begin{tabular}[c]{@{}c@{}} 51.2 / 43.2 \\  93.9 / 97.3 \\ -- / -- \\ 75.7 / 30.7 \\ 50.0 / 38.0 \\ 68.7 / 38.7 \end{tabular} &
\begin{tabular}[c]{@{}c@{}} Xilinx Virtex-5 \\  Altera 5SGSMD5K2 \\ AItera Cyclone IV  \\ Xilinx ZC706 \\ AItera Cyclone IV GX, Stratix IV GX \\ Xilinx Ultrascale + ZCU102 \end{tabular} \\
\hline
\begin{tabular}[c]{@{}c@{}} Efficient Large-Scale \\  Stereo Matching \end{tabular} & 
\begin{tabular}[c]{@{}c@{}} Rahnama et al.~\cite{rahnama2018real} \\  Rahnama et al.~\cite{rahnama2019real} \end{tabular} & 
\begin{tabular}[c]{@{}c@{}} 47 \\  50 \end{tabular} & 
\begin{tabular}[c]{@{}c@{}} 1242 $\times$ 375 \\  1242 $\times$ 375 \end{tabular} & 
\begin{tabular}[c]{@{}c@{}} -- \\  -- \end{tabular} & 
\begin{tabular}[c]{@{}c@{}} -- \\  -- \end{tabular} & 
\begin{tabular}[c]{@{}c@{}} 2.91\\  5 \end{tabular} &
\begin{tabular}[c]{@{}c@{}} 11.9 / 15.7 \\  70.7 / 8.7 \end{tabular} &
\begin{tabular}[c]{@{}c@{}} Xilinx ZC706 \\  Xilinx ZCU104 \end{tabular} \\
\bottomrule[1.5pt]
\end{tabular}%
\caption{Comparison of Stereo Vision Systems on FPGA platforms, across local stereo matching, global stereo matching, semi-global stereo matching (SGM) and efficient large-scale stereo matching (ELAS) algorithms. The results reported in each design are evaluated by frame rate (fps), image resolution (width $\times$ height), disparity levels, million disparity estimations per second (MDE/s), power (W), resource utilization (logic\% and BRAM\%) and hardware platforms, where MDE/s = width $\times$ height $\times$ fps $\times$ disparity.}
\label{tab:perception-table}
}
\end{table*}

\subsection{Semi-Global Matching on FPGA}
Semi-global matching (SGM)~\cite{hirschmuller2005accurate} bridges the gap between local and global methods, and achieves a notable improvement in accuracy. SGM calculates the initial matching disparities by comparing local pixels, and then approximates an image-wide smoothness constraint with global optimization, which can obtain more robust disparity maps through this combination. There are several critical challenges for implementing SGM on hardware, e.g., data dependence, high complexity, and large storage, so this is an active research field with recent works proposing FPGA-friendly variants of SGM~\cite{wang2015real,honegger2014real,banz2010real,mattoccia2015passive, gehrig2009real}.

Banz et al.~\cite{banz2010real} propose a systolic-array based hardware architecture for SGM disparity estimation along with a two-dimensional parallelization concept for SGM. This design achieves 30 fps performance at 640×480 pixel images with a 128-disparity range on the Xilinx Virtex-5 FPGA platform. Wang et al.~\cite{wang2015real} implement a complete real-time FPGA-based hardware system that supports both absolute difference-census cost initialization, cross-based cost aggregation and semi-global optimization. The system achieves 67 fps at 1024×768 resolution with 96 disparity levels on the Altera Stratix-IV FPGA platform, and 42fps at 1600×1200 resolution with 128 disparity levels on the Altera Stratix-V FPGA platform. The design in Cambuim et al.~\cite{cambuim2017hardware} uses a scalable systolic-array based architecture for SGM based on the Cyclone IV FPGA platform, and it achieves a 127 fps image delivering rate in 1024×768 pixel HD resolution with 128 disparity levels. The key point of this design is the combination of disparity and multi-level parallelisms such as image line processing to deal with data dependency and irregular data access pattern problems in SGM. Later, to improve the robustness of SGM and achieve a more accurate stereo matching, Cambuim et al.~\cite{cambuim2019fpga} combine the sampling-insensitive absolute difference in the pre-processing phase, and propose a novel streaming architecture to detect noisy and occluded regions in the post-processing phase. The design is evaluated in a full stereo vision system using two heterogeneous platforms, DE2i-150 and DE4, and achieves a 25 fps processing rate in 1024×768 HD maps with 256 disparity levels. 


While most existing SGM designs on FPGA are implemented using the register-transfer level (RTL), some works leveraged the high-level synthesis (HLS) approach. Rahnama et al.~\cite{rahnama2018r3sgm} implement an SGM variation on FPGA using HLS, which achieves 72 fps speed at 1242×375 pixel size with 128 disparity levels. To reduce the design effort and achieve an appropriate balance among speed, accuracy and hardware cost, Zhao et al.~\cite{zhao2020fp} recently propose FP-Stereo for building high-performance SGM pipelines on FPGAs automatically. A series of optimization techniques are applied in this system to exploit parallelism and reduce resource consumption. Compared to GPU designs~\cite{hernandez2016embedded}, it achieves the same accuracy at a competitive speed while consuming much less energy.

\begin{figure*}[t!]
        \centering\includegraphics[width=\columnwidth]{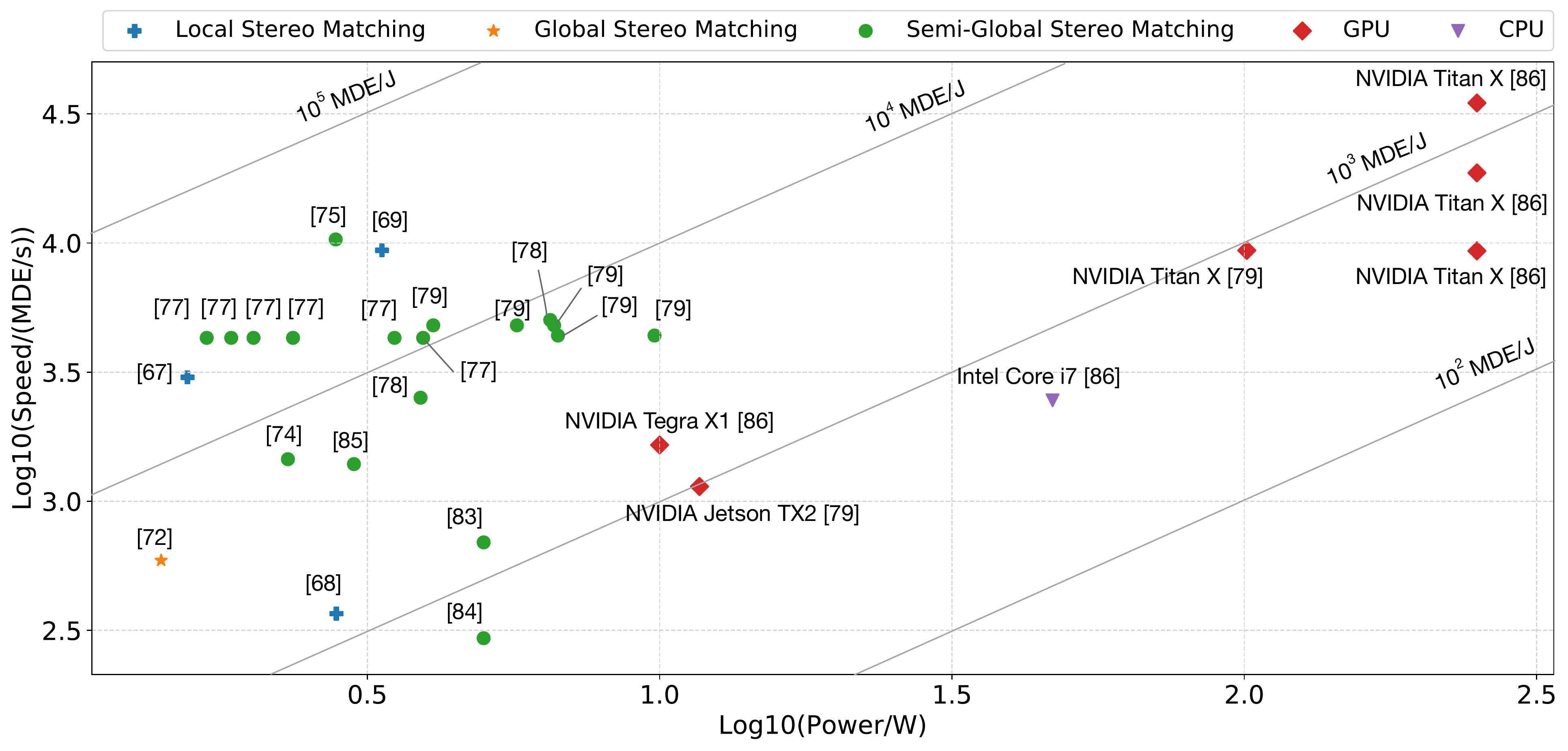}
        \caption{A comparison between different designs for perception tasks on a logarithm coordinate of power (W) and performance (MDE/s).}
        \label{fig:power_MDE}
\end{figure*}

To compare these implementations, the depth quality of are evaluated on Middlebury Benchmark~\cite{hirschmuller2007evaluation}, with four image pairs \textit{Tsukuba}, \textit{Venus}, \textit{Teddy}, \textit{Cones}. As shown in Tab.~\ref{tab:MDE_accuracy}, there is a general trade-off between accuracy and processing speed. The stereo vision system designs in Tab.~\ref{tab:perception-table} are drawn as points in Fig.~\ref{fig:power_MDE} (if both power and speed number are reported), using $log_{10}(power)$ as \textit{x}-coordinate and $log_{10}(speed)$ as \textit{y}-coordinate ($y$-$x$=$log_{10}(energy\_efficiency)$). Besides FPGA-based implementations, we also plot GPU and CPU experimental results as a comparison to FPGA designs' performance. In general, local and semi-global stereo matching designs have achieved higher performance and energy efficiency than global stereo matching designs. As introduced in section~\ref{subsec:global_SM}, global stereo matching algorithms usually involve massive computational-intensive optimization techniques. Even for the same design, varying design parameters (e.g., window size) may result in a 10× difference in energy efficiency. Compared to GPU and CPU-based designs, FPGA-based designs have achieved higher energy efficiency, and the speed of many FPGA implementations have surpassed general-purpose processors.

\begin{table*}[t!]
\centering
\renewcommand*{\arraystretch}{1.2}
\resizebox{\columnwidth}{!}{%
\small
\begin{threeparttable}
\begin{tabular}{c|c|c|c|c|c|c|c|c|c|c|c|c|c|c}
\toprule[1.5pt]
\multirow{2}{*}{\textbf{Reference}} & \multirow{2}{*}{\textbf{MDE/s}} & \multicolumn{3}{c|}{\textbf{Tsukuba}}          & \multicolumn{3}{c|}{\textbf{Venus}}            & \multicolumn{3}{c|}{\textbf{Teddy}}            & \multicolumn{3}{c|}{\textbf{Cones}}             & \multirow{2}{*}{\textbf{\begin{tabular}[c]{@{}c@{}}Average Bad\\ Pixel Rate\end{tabular}}} \\ \cline{3-14}
                                    &                                 & \textbf{nonocc}\tnote{1} & \textbf{all}\tnote{2} & \textbf{disc}\tnote{3} & \textbf{nonocc} & \textbf{all} & \textbf{disc} & \textbf{nonocc} & \textbf{all} & \textbf{disc} & \textbf{nonocc} & \textbf{all} & \textbf{disc} &                                                                                            \\ \hline
Shan et al.~\cite{shan2012fpga}                             & 15437                           & -               & 24.5         & -             & -               & 15.7         & -             & -               & 15.1         & -             & -               & 14.1          & -             & all = 17.3                                                                                       \\ \hline
Shan et al.~\cite{shan2014hardware}                           & 13076                           & 3.62            & 4.15         & 14.0          & 0.48            & 0.87         & 2.79          & 7.54            & 14.7         & 19.4          & 3.51            & 11.1          & 9.64          & 7.65                                                                                       \\ \hline
Wang et al.~\cite{wang2015real}                             & 10472                           & 2.39            & 3.27         & 8.87          & 0.38            & 0.89         & 1.92          & 6.08            & 12.1         & 15.4          & 2.12            & 7.74          & 6.19          & 5.61                                                                                       \\ \hline
Jin et al.~\cite{jin2014fast}                              & 9362                            & 1.66            & 2.17         & 7.64          & 0.4             & 0.6          & 1.95          & 6.79            & 12.4         & 17.1          & 3.34            & 8.97          & 9.62          & 6.05                                                                                       \\ \hline
Jin et al.~\cite{jin2009fpga}                              & 4522                            & 9.79            & 11.6         & 20.3          & 3.59            & 5.27         & 36.8          & 12.5            & 21.5         & 30.6          & 7.34            & 17.6          & 21.0          & 17.2                                                                                       \\ \hline
Zhang et al.~\cite{zhang2011real}                            & 3020                            & 3.84            & 4.34         & 14.2          & 1.2             & 1.68         & 5.62          & 7.17            & 12.6         & 17.4          & 5.41            & 11.0          & 13.9          & 8.2                                                                                        \\ \hline
Banz et al.~\cite{banz2010real}                              & 1455                            & 4.1             & -            & -             & 2.7             & -            & -             & 11.4            & -            & -             & 8.4             & -             & -             & nonocc = 6.7                                                                                 \\ \hline
Jin et al.~\cite{jin2012real}                             & 590                            & 1.43            & 2.51         & 6.6           & 2.37            & 2.97         & 13.1          & 8.11            & 13.6         & 15.5          & 8.12            & 13.8          & 16.4          & 8.71                                                                                       \\ 
\bottomrule[1.5pt]
\end{tabular}
\begin{tablenotes}
            \item[1] nonocc: average percentage of bad pixels in non-occluded regions.
            \item[2] all: average percentage of bad pixels in all regions.
            \item[3] disc: average percentage of bad pixels in discontinuous regions.
        \end{tablenotes}
\end{threeparttable}
\caption{A comparison between different designs on performance (MDE/s) and accuracy results on Middlebury Benchmark. (The lower of average bad pixel rate means the better stereo matching performance)}
\label{tab:MDE_accuracy}
}
\end{table*}

\subsection{Efficient Large-Scale Stereo Matching on FPGA}
Another popular stereo matching algorithm that offers a good trade-off between speed and accuracy is Efficient Large-Scale Stereo Matching (ELAS)~\cite{geiger2010efficient}, which is currently one of the fastest and accurate CPU algorithms concerning the resolution on Middlebury dataset. ELAS implements a slanted plane prior very effectively while its dense estimation of depth is completely decomposable over all pixels, which make it attractive for easily parallelized.

Rahnama et al.~\cite{rahnama2018real} first implement and evaluate an FPGA accelerated adaptation of the ELAS algorithm, which achieved a frame rate of 47 fps (up to 30× compared high-end CPU) while consuming under 4W of power. By taking advantage of different components on the SoC, several elaboration blocks such as feature extraction and dense matching are executed on FPGA, while I/O and other conditional/sequential blocks are executed on ARM-core CPU. The authors also reveal the strategy to accelerate complex and computationally diverse algorithms for low power and real-time systems by collaboratively utilizing different compute components. Later, by leveraging and combining the best features of SGM and ELAS-based methods, Rahnama et al.~\cite{rahnama2019real} propose a sophisticated stereo approach and achieve an 8.7\% error rate on the challenging KITTI 2015 dataset at over 50 fps, with a power consumption of only 4.5 W.

\subsection{CNN-based stereo vision system on FPGA}
Convolutional neural networks (CNNs) have been demonstrated to perform very well on many vision tasks such as image classification, object detection, and semantic segmentation. Recently, CNN has also been utilized in stereo estimation~\cite{zagoruyko2015learning,vzbontar2016stereo} and stereo matching~\cite{luo2016efficient}. CNN is applied to determine SGM penalties~\cite{seki2017sgm}, estimate real-time optical flow disparity~\cite{mayer2016large} and predict cost volume computation and aggregation~\cite{kuzmin2017end}. 

CNN has been deployed on FPGA platforms in several works~\cite{li2016high,qiu2016going,guo2017angel,yu2018instruction}, with an example of lightweight YOLOv2 for object detection~\cite{nakahara2018lightweight}. Nakahara et al. implement a pipelined-based architecture for lightweight YOLOv2 with a binarized CNN on Xilinx ZCU102 FPGA platform. This design achieves a 40.81 fps object detection speed, which is 177.4× faster than ARM Cortex-A57 and 27.5× faster than NVIDIA Pascal embedded GPU. Many FPGA-based CNN accelerator implementations have been summarized in~\cite{guo2019dl}.

\section{Localization on FPGA}
\label{sec:localization}

\subsection{Overview}
For robots, one of the most critical tasks is localization and mapping. Simultaneous Localization and Mapping (SLAM) is an advanced robot navigation algorithm for constructing or updating a map of unknown surroundings while simultaneously keeping tracking the robot’s location. Localization and mapping are two concurrent tasks and cannot be solved independently from each other. Localizing a robot requires a sufficiently detailed map, and constructing or updating or a map requires accurate landmarks or pose estimates from known positions. 

Many SLAM algorithms have been developed in the last decades to improve the accuracy and robustness, and its implementation comes in a diverse set of sizes and shapes. One end of the spectrum is dense SLAM algorithms \cite{belshaw2008high,williams2017evaluation,van2019fpga,gautier2014real}, which can generate high-quality maps of the environment with complex computations. Dense SLAM algorithms usually are executed on powerful and high-performance machines to ensure real-time performance. At the same time, the intensive computation characteristic makes dense SLAM hard to deploy on edge devices. The other end of the spectrum is sparse SLAM~\cite{bailey2006consistency,mur2015orb,montemerlo2002fastslam,gu2015fpga}, which is computationally light by only selecting limited numbers of landmarks or features. 

To form a compromise in terms of compute intensity and accuracy quality between these two extremes, a family of works described as semi-dense SLAM has emerged~\cite{cadena2016past,engel2013semi}. They aim to achieve better computational efficiency compared to dense methods by only processing a subset of high-quality sensory information while providing a more dense and informative map compared to sparse methods. 

A typical SLAM system includes two components: the front-end and the back-end, which are with different computational characteristics. The front-end associates sensory measurements in consecutive frames to physical landmarks. It incrementally deduces the robot motion by applying geometry constraints on the associated sensory observations. The back-end tries to minimize errors introduced from sensory measurement noises by performing optimizations on a batch of observed landmarks and tracked poses. Filter based (e.g., Extended Kalman Filter) and numerical optimization based (e.g., bundle adjustment) algorithms are two prevalent methods for SLAM back-end. 

A critical challenge to mobile robot localization is accuracy and efficiency under stringent power and resource constraints. To avoid losing tracked features due to large motions between consecutive frames, SLAM systems need to process sensory data at a high frame rate. For example, open data sets for evaluating localization algorithms \cite{geiger2013vision,burri2016euroc} for drones and vehicles provide images at 10 to 20 fps. Low power computing systems are always required to extend the battery life of mobile robots.     
Most SLAM algorithms are developed on CPU or GPU platforms, of which power consumption is hundreds of Watts. To execute SLAM efficiently on mobile robots and meet real-time and power constraints, specialized chips and accelerators have been developed. FPGA SoCs provide rich sensor interfaces, dedicated hardware logic and programmability, hence they have been explored in diverse ways in recent years. We summarize and discuss FPGA-based accelerators for SLAMs in the following sections. 

\subsection{Dense SLAM on FPGA}
Dense SLAM can construct high quality and complete models of the environment, and most of them are running in high-end hardware platforms (especially GPU). 
One of the representative real-time dense SLAM algorithms is KinectFusion~\cite{newcombe2011kinectfusion}, which was released by Microsoft in 2011. As a scene reconstruction algorithm, it continuously updates the global 3D map and tracks the location of depth cameras within the surrounding environment. KinectFusion is generally composed of three algorithms: ray-casting algorithm for generating graphics from surface information, iterative closest point (ICP) algorithm for camera-tracking and volumetric integration (VI) algorithm for integrating depth streams into the 3D surface. Several works have attempted to implement real-time dense SLAM algorithms on a heterogeneous system with FPGA embedded. 

Several works implement computationally intensive components of dense SLAMs, such as ICP and VI, on FPGA to accelerate the critical path.
Belshaw~\cite{belshaw2008high} presents an FPGA implementation of the ICP algorithm, which achieves over 200 fps tracking speed with low tracking errors. This design divides the ICP algorithm into filtering, nearest neighbor, transform recovery and transform application stages. It leverages fixed-point arithmetic and the power of two data points to utilize FPFA logic efficiently. Williams~\cite{williams2017evaluation} notices that the nearest neighbor search takes up the majority of ICP runtime, and then proposes two hybrid CPU-FPGA architectures to accelerate the bottleneck of the ICP-SLAM algorithm. The implementation is performed with Vivado HLS, a high-level synthesis tool from Xilinx, and achieves a maximum 17.22× speedup over the ARM software implementation. Hoorick~\cite{van2019fpga} presents an FPGA-based heterogeneous framework using a similar HLS method to accelerate the KinectFusion algorithm and explored various ways of dataflow and data management patterns. Gautier et al.~\cite{gautier2014real} embed both ICP and VI algorithms on an Altera Stratix V FPGA by using the OpenCL language and the Altera OpenCL SDK. This design was a heterogeneous system with NVIDIA GTX 760 GPU and Altera Stratix V FPGA. By distributing different workloads on different parts of SoC, the entire system achieves up to 28 fps real-time speed.

\subsection{Sparse SLAM on FPGA}
Sparse SLAM algorithms usually use a small set of features to track and maintain a sparse map of surrounding environments. These algorithms exhibit lower power consumption but are limited to the localization accuracy. 

\subsubsection{EKF-SLAM}
EKF-SLAM~\cite{bailey2006consistency} is a class of algorithms that utilizes the extended Kalman Filter (EKF) for SLAM. EKF-SLAM algorithms are typically feature-based and use the maximum likelihood algorithm for data association. Several heterogeneous architectures using multi-core CPUs, GPUs, DSPs, and FPGAs are proposed to accelerate the complex computation in EKF-SLAM algorithms. Bonato et al.~\cite{bonato2009floating} presents the first FPGA-based architecture for the EKF-SLAM based algorithm that is capable of processing 2D maps at up to 1800 features at real-time with a frequency of 14 Hz, compared to 572 features with Pentium CPU and 131 features with ARM. They analyze the computational complexity and memory bandwidth requirements for FPGA-based EKF-SLAM, and then propose an architecture with a parallel memory access pattern to accelerate the matrix multiplication. This design achieves two orders of magnitude more power-efficient than a general-purpose processor.

Similarly, Tertei et al.~\cite{tertei2014fpga} propose an efficient FPGA-SoC hardware architecture for matrix multiplication with systolic arrays to accelerate EKF-SLAM algorithms. The setup of this design is a PLB peripheral to PPC440 hardcore embedded processor on a Virtex5 FPGA, and it achieves a 7.3× speedup with a processing frequency of 44 Hz compared to the pure software implementation. Later, taking into account the symmetry in cross-covariance matrix-related computations, Tertei et al.~\cite{tertei2016fpga} improve the previous implementation to further reduce the computational time and on-chip memory storage on Zynq-7020 FPGA. 

DSP is also leveraged in some works to accelerate EKF-SLAM algorithms. Vincke et al.~\cite{vincke2012real} implement an efficient implementation of EKF-SLAM on a low-cost heterogeneous architecture system consisting of a single-core ARM processor with a SIMD coprocessor and a DSP core. The EKF-SLAM program is partitioned into different functional blocks based on the profiling characteristics results. Compared to a non-optimized ARM implementation, this design achieved 4.7× speed up from 12 fps to 57 fps. In a later work, Vincke et al.~\cite{vincke2014simd} replace the single-core ARM with a double-core ARM to optimize the non-optimized blocks using the OpenMP library. This design achieves a 2.75× speedup compared to non-optimized implementation.

\subsubsection{ORB-SLAM}
ORB-SLAM~\cite{mur2015orb} is an accurate and widely-used sparse SLAM algorithm for monocular, stereo, and RGB-D cameras. Its framework usually consists of five main procedures: feature extraction, feature matching, pose estimation, pose optimization and map updating. Based on the profiling results on a quad-core ARM v8 mobile SoC, feature extraction is the most computation-intensive stage in the ORB-SLAM system, which consumes more than half of CPU resources and energy budget~\cite{fang2017fpga}. 

ORB based feature extraction algorithm usually consists of two parts, namely Oriented Feature from Accelerated Segment Test (oFAST)~\cite{biadgie2014feature} based feature detection and Binary Robust Independent Elementary (BRIEF)~\cite{calonder2010brief} based feature descriptors computation. To accelerate this bottleneck, Fang et al.~\cite{fang2017fpga} design and implement a hardware ORB feature extractor and achieved a great balance between performance and energy consumption, which outperforms ARM Krait by 51\% and Intel Core i5 by 41\% in computation latency as well as outperforms ARM Krait by 10\% and Intel Core i5 by 83\% in energy consumption. Liu et al.~\cite{liu2019eslam} propose an energy-efficient FPGA implementation eSLAM to accelerate both feature extraction and feature matching stages. This design achieves up to 3× and 31× speedup in framerate, as well as up to 71× and 25× in energy efficiency improvement compared to Intel i7 and ARM Cortex-A9 CPUs, respectively. This eSLAM design utilizes a rotationally symmetric ORB descriptor pattern to make the algorithm more hardware-friendly, resulting in a 39\% less latency compared to~\cite{fang2017fpga}. Rescheduling and parallelizing optimization techniques are also exploited to improve the computation throughput in eSLAM design. 

Scale-invariant feature transform (SIFT) and Harris corner detector are also commonly-used feature extraction methods. SIFT is invariant to rotation and translation. Gu et al.~\cite{gu2015fpga} implement SIFT-feature based SLAM algorithm on FPGA and accelerate the matrix computation part to achieve speedup. Harris corner detector is used to extract corners and features of an image, and Schulz et al.~\cite{schulz2016harris} propose an implementation of Harris and Stephen corner detector optimized for an embedded SoC platform that integrates a multicore ARM processor with Zynq-7000 FPGA. Taking into account I/O requirements and the advantage of parallelization and pipeline, this design achieves a speedup of 1.77 compared to dual-core ARM processors.


\subsubsection{Fast-SLAM}
One of the key limitations of EKF-SLAM is its computational complexity since EKF-SLAM requires time quadratic in the number of landmarks to incorporate each sensor update. In 2002, Montemerlo et al.~\cite{montemerlo2002fastslam} propose an efficient SLAM algorithm called Fast-SLAM. Fast-SLAM decomposes the SLAM problem into a robot localization problem and a landmark estimation problem. It recursively estimates the full posterior distribution over landmark positions and robot path with a logarithmic scale. 

Abouzahir et al.~\cite{abouzahir2016large} implement Fast-SLAM 2.0 on a CPU-GPGPU-based SoC architecture. The algorithm is partitioned into function blocks, and each of them is implemented on the CPU or GPU accordingly. This optimized and efficient CPU-GPGPU partitioning enables accurate localization and a 37× execution speedup compared to non-optimized implementation on a single-core CPU. Further, Abouzahir et al.~\cite{abouzahir2018embedding} perform a complete study of the processing time of different SLAM algorithms under popular embedded devices, and demonstrate that Fast-SLAM2.0 allowed a compromise between the consistency of localization results and computation time. This algorithm is then optimized and implemented on GPU and FPGA using HLS and parallel computing frameworks OpenCL and OpenGL. It is observed that the global processing time of FastSLAM2.0 on FPGA implementations achieves 7.5× acceleration compared to high-end GPU. The processing frequency achieves 102 fps and meets the real-time performance constraints of an operated robot.

\subsubsection{VO-SLAM}
The visual odometry based SLAM algorithm (VO-SLAM) also belongs to the Sparse SLAM class with low computational complexity. Gu et al.~\cite{gu2015fpga} implement the VO-SLAM algorithm on a DE3 board (Altera Stratix III) to perform drift-free pose estimation, resulting in localization results accurate to 1-2cm. A Nios II soft-core is used as a master processor. The authors design a dedicated matrix accelerator and propose a hierarchical matrix computing mechanism to support application requirements. This design achieves a processing speed of 31 fps with 30000 global map features, and 10× energy saving for each frame processing compared to Intel i7 CPU.

\subsection{Semi-dense SLAM on FPGA}
Semi-dense SLAM algorithms have emerged to provide a compromise between sparse SLAM and dense SLAM algorithms, which attempt to achieve improved efficiency and dense point clouds. However, they are still usually computationally intensive and require multicore CPUs for real-time processing.

Large-Scale Direct Monocular SLAM (LSD-SLAM) is one of the state-of-the-art and widely-used semi-dense SLAM algorithms, and it directly operates on image intensities for both tracking and mapping problems. The camera is tracked by direct image alignment, while geometry is estimated from semi-dense depth maps acquired by filtering over multiple stereo pixel-wise comparisons. 

Several works have explored LSD-SLAM FPGA-SoC implementation. Boikos et al.~\cite{boikos2016semi} investigate the performance and acceleration opportunities for LSD-SLAM in the SoC system. This design achieves an average framerate of more than 4 fps for a resolution of 320×240 with an estimated power of less than 1W, which is a 2× acceleration and more than 4.3× energy efficiency compared to a software version running on embedded CPUs. The author also notes that the communication between two accelerators is via DDR memory since the produced intermediate data is too large to be fully cached on the FPGA. Hence, it is important to optimize the memory architecture (e.g., data movement and caching techniques) to ensure the scalability and compatibility of the design. 

To further improve the performance of~\cite{boikos2016semi}, Boikos et al.~\cite{boikos2017high} re-implement the design using a dataflow architecture and distributed asynchronous blocks to allow the memory system and the custom hardware pipelines to function at peak efficiency. This implementation can process and track more than 22 fps with an embedded power budget and achieves a 5× speedup over~\cite{boikos2016semi}. 

Furthermore, Boikos et al.~\cite{boikos2019scalable} combine a scalable depth estimation with direct semi-dense SLAM architecture and propose a complete accelerator for semi-dense SLAM on FPGA. This architecture achieved more than 60 fps at the resolution of 640×480 and an order of magnitude power consumption improvement compared to Intel i7-4770 CPU. This implementation leverages multi-rate and multi-modal units to deal with LSD-SLAM’s complex control flow. A new dataflow paradigm is also proposed where the kernel is linked with a single consumer and a single producer to achieve high efficiency. 

\subsection{CNN-based SLAM}
Recently, CNNs have made significant progress in the perception and localization ability of the robots compared to handcrafted methods. Take one of the main SLAM components, feature extraction, as an example, the CNN-based approach SuperPoint~\cite{detone2018superpoint} can achieve 10\%-30\% higher matching accuracy compared to handcrafted ORB. Other CNN-based methods, such as DeepDesc~\cite{simo2015discriminative} and GeM~\cite{radenovic2018fine}, also present significant improvements in feature extraction and descriptor generation stage. However, CNN has a much higher computational complexity and requires more memory footprint. 

Several works have explored to deploy CNN on FPGAs. Xilinx DPU~\cite{knuthwebsite} is one of the state-of-the-art programmable engines dedicated to CNN, which has a specialized instruction set and works efficiently across various CNN topologies. Xu et al.~\cite{xu2020cnn} propose a hardware architecture to accelerate CNN-based feature extraction SuperPoint on the Xilinx ZCU102 platform and achieve 20 fps in a real-time SLAM system. The key point of this design is an optimized software dataflow to deal with the extra post-processing operations within CNN-based feature extraction networks. 8-bit fixed-point numerics are leveraged in the post-processing operations and CNN backbone. Similar hardware-oriented model compression techniques (e.g., data quantization and weight reduction) have been widely adopted in robotics and CNN related designs~\cite{han2015deep,krishnan2019quantized,langroudi2020adaptive,tambe2020algorithm,li2016ternary,choi2019accurate,kim2020position,tambe2019adaptivfloat}.

Yu et al.~\cite{yu2020cnn} build a CNN-based monocular decentralized-SLAM (DSLAM) on the Xilinx ZCU102 MPSoC platform with DPU. DSLAM is usually used in multi-robot applications that can share environment information and locations between agents. 
To accelerate the main components in DSLAM, namely visual odometry (VO) and decentralized place recognition (DPR), the authors adopt CNN-based Depth-VO-Feat~\cite{zhan2018unsupervised} and NetVLAD~\cite{arandjelovic2016netvlad} to replace handcrafted approaches and propose a cross-component pipeline scheduling algorithm to improve the performance. 

To enable multi-tasking processing in embedded robots on CNN accelerators, Yu et al.~\cite{yuinca} further propose an INterruptible CNN accelerator (INCA) with a novel virtual-instruction-based interrupt method. Feature extraction and place recognition of DSLAM are deployed and accelerated on the same CNN accelerator of the embedded FPGA system, and the interrupt response latency is reduced by 1\%.


\subsection{Bundle Adjustment}
Besides the hardware implementation of the frontend of the SLAM system, several works investigate to accelerate the backend of the SLAM system, mainly Bundle Adjustment (BA). BA is heavily used in robot localization~\cite{mur2015orb,mur2017orb}, autonomous driving~\cite{liu2020eavr}, space exploration missions~\cite{maimone2007two} and some commercial products~\cite{klingner2013street}, where it is usually employed in the last stage of the processing pipeline to refine camera trajectories and 3D structures further. 

Essentially, BA is a massive joint non-linear optimization problem that usually consumes a significant amount of power and processing time in both offline visual reconstruction and real-time localization applications. 

Several works aim to accelerate BA on multi-core CPUs or GPUs using parallel or distributed computing techniques. Jeong et al.~\cite{jeong2011pushing} exploit efficient memory handling and fast block-based linear solving, and propose a novel embedded point iterations method, which substantially improves the BA performance on CPU. Wu et al.~\cite{wu2011multicore} present a multi-core parallel processing solution running on CPUs and GPUs. The matrix-vector product is carefully restructured in this design to reduce memory requirements and compute latency substantially. Eriksson et al.~\cite{eriksson2016consensus} propose a distributed approach for very large scale global bundle adjustment computation to achieve BA performance improvement. The authors present a consensus framework using the proximal splitting method to reduce the computational cost. Similarly, Zhang et al.~\cite{zhang2017distributed} propose a distributed formulation to accelerate the global BA computation without much distributed computing communication overhead. 

To better deploy BA in embedded systems with strict power and real-time constraints, recent works explore BA algorithm acceleration using specialized hardware. The design in~\cite{suleiman2019navion} implements both the image frontend and BA backend of a VIO algorithm on a single-chip for nano-drone scale applications. Liu et al.~\cite{liu2020pi} propose a hardware-software co-designed BA hardware accelerator and its implementation on an embedded FPGA-SoC to achieve higher performance and power efficiency simultaneously. Especially, a co-observation optimization technique and a hardware-friendly differentiation method are proposed to accelerate BA operations with optimized usage of memory and computation resources. Sun et al.~\cite{sun2020bax} present a hardware architecture running local BA on FPGAs, which works without external memory access and refines both cameras poses and 3D map points simultaneously.

\subsection{Discussion}
We summarize FPGA based SLAM systems in Tab.~\ref{tab:localization-table}. 
It only includes works that implement the whole SLAM on an FPGA and provide overall performance and power evaluation.
The works in the table adopt a similar FPGA-SoC architecture that accelerates computationally intensive components by FPGA fabrics and offloads others works to embedded processors on FPGAs. Compared with sparse method, the semi-dense implementation has lower frame rate, which is mainly due to the high resolution data processed in the pipeline. Due to the high frame rates and low power consumption, sparse SLAM FPGA have been used in drones and autonomous vehicles \cite{yu2020micro}. The two sparse SLAM implementations achieve similar performance in terms of frame rate. Compared with the ORB design, the VO SLAM design includes pre-processing and outliers removal hardware, such as image rectification and RANSAC, which lead to a more accurate but power inefficient implementation.


\begin{table}[t!]
\centering
\renewcommand*{\arraystretch}{1}
\resizebox{\columnwidth}{!}{%
\begin{tabular}{c|c|c|c|c|c}
\toprule[1.2pt]
                                                         & \textbf{Method}                                               & \textbf{Platform}                                                      & \begin{tabular}[c]{@{}c@{}}\textbf{Frame} \\ \textbf{Rate}\end{tabular} & \textbf{Power} & \begin{tabular}[c]{@{}c@{}}\textbf{Indoor}\\ \textbf{Error}\end{tabular} \\ \hline
\begin{tabular}[c]{@{}c@{}}Boikos \\ et al.~\cite{boikos2016semi}\end{tabular} & \begin{tabular}[c]{@{}c@{}}Semi\\ Dense\end{tabular} & \begin{tabular}[c]{@{}c@{}}Xilinx Zynq\\ 7020 SoC\end{tabular} & 4.5 fps                                                & 2.5 W  & na                                                     \\ \hline
\begin{tabular}[c]{@{}c@{}}Liu \\ et al.~\cite{liu2019eslam}\end{tabular}    & ORB                                                  & \begin{tabular}[c]{@{}c@{}}Xilinx Zynq\\ 7000 SoC\end{tabular} & 31 fps                                                 & 1.9 W  & 4.5 cm                                                  \\ \hline
\begin{tabular}[c]{@{}c@{}}Gu \\ et al.~\cite{gu2015fpga}\end{tabular}     & VO                                                   & \begin{tabular}[c]{@{}c@{}}Altera \\ Stratix III\end{tabular}  & 31 fps                                                 & 5.9 W  & 2 cm                                                    \\
\bottomrule[1.2pt]
\end{tabular}%
\caption{Comparison of FPGA SLAM Systems.}
\label{tab:localization-table}
}
\end{table}

\section{Planning and Control on FPGA}
\label{sec:planning_control}
\subsection{Overview}
Planning and control are the modules that compute how the robot should maneuver itself. They usually include behavioral decision, motion planning and feedback control kernels. Without loss of generality, we focus on the motion planning algorithms and their FPGA implementations in this section.

As a fundamental problem in the robotic system, motion planning aims to find the optimal collision-free path from the current position to a goal position for a robot in complex surroundings. Generally, motion planning contains three steps, namely roadmap construction, collision detection and graph search~\cite{kavraki1996probabilistic,leven2002framework}. Motion planning will become a relatively complicated problem when robots work with a high degree of freedom (DOF) configurations since the search space will be exponentially increased. Typically, state-of-the-art CPU-based approaches take a few seconds to find a collision-free trajectory~\cite{karaman2011sampling,gammell2015batch,hauser2015lazy}, making the existing motion planning algorithms too slow to meet the real-time requirement for complex robot tasks and environments. Several works have investigated approaches to speed up motion planning, either for each stage or whole pipeline.

\subsection{Roadmap Construction}
In the roadmap construction step, the planner generates a set of states in the robot’s configuration space and then connects them with edges to construct a general-purpose roadmap in the obstacle-free space. Each state represents a robot’s configuration, and each edge represents a possible robot movement. Conventional algorithms build the roadmap by randomly sampling poses from configuration space at runtime to navigate around the obstacles present at that time. 

Several works explore roadmap construction acceleration. Yershova et al.~\cite{yershova2007improving} improve the nearest neighbor search to accelerate roadmap construction by orders of magnitude compared to the naive nearest-neighbor searching. Wang et al.~\cite{wang2015fast} reduce the computation workload by trimming roadmap edges and keeping the roadmap to a reasonable size to achieve speedup. Different from online runtime approaches, Murray et al.~\cite{murray2019programmable} completely remove the runtime latency by conducting the roadmap construction only once at the design time. A more general and much larger roadmap is precomputed and allows for fast and successive queries in complex environments without reprogramming the accelerator during runtime.

\subsection{Collision Detection}
In the collision detection step, the planner determines whether there are potential collisions with the environment or the robot itself during movement. Specifically, collision detection is the primary challenge in motion planning, which often comprises 90\% of the processing time~\cite{bialkowski2011massively}. 

Several works leverage data parallelization computing on GPUs to achieve speedup~\cite{pan2012gpu,pan2010g,bialkowski2011massively}. For example, Bialkowski et al.~\cite{bialkowski2011massively} divide the RRT* algorithm of collision detection tasks into three parallel dimensions and construct thread block grids to execute collision computations simultaneously. However, GPU can only provide a constant speedup factor due to the core limitations, which is still hard to achieve the real-time requirement. 

Recently, \cite{atay2006motion,murray2016microarchitecture,lian2018dadu} develop high-efficiency custom hardware implementations based on the FPGA system. Atay and Bayazit~\cite{atay2006motion} focus on directly accelerating the PRM algorithm on FPGA by creating functional units to perform the random sampling, nearest neighbor search and parallelizing triangle-triangle testing. However, this design cannot be reconfigured at runtime, and the huge resource demands make it fail to support a large roadmap. Murray et al.~\cite{murray2016microarchitecture} present a novel microarchitecture for an FPGA-based accelerator to speed up collision detection by creating a specialized circuit for each motion in the roadmap. This solution achieves sub-millisecond speed for motion planning query and improves the power consumption by more than one order of magnitude, which is sufficient to enable real-time robotics applications.

Besides real-time constraint, motion planning algorithms also have flexibility requirements to make the robots adapt to dynamic environments. Dadu-P~\cite{lian2018dadu} build a scalable motion planning accelerator to attain both high efficiency and flexibility, where a motion plan can be solved in around 300 microseconds in a dynamic environment. A hardware-friendly data structure representing roadmap edges is adopted to achieve flexibility, and a batched processing as well as a priority-rating method are proposed to achieve high efficiency. But this design comprises a 25× latency increase to make it retargetable to different robots and scenarios due to the external memory access. Murray et al.~\cite{murray2019programmable} develop a fully retargetable microarchitecture of collision detection and graph search accelerator that can perform motion planning in less than 3 ms with a modest power consumption of 35 W. This design divides the collision detection workflow into two stages. The collision detection results for the discretized roadmap are precomputed in the first stage before runtime, and then the collision detection accelerator streams in the voxels of obstacles and the edges of flags which are in collision at runtime.

\subsection{Graph Search}
After collision detection, the planner will try to find the shortest and safe path from the start position to the target position based on the obtained collision-free roadmap through graph search. Several works explore graph search accelerations. Bondhugula et al.~\cite{bondhugula2006hardware} employ a parallel FPGA-based design using a blocked algorithm to solve large instances of All-Pairs Shortest-Paths (APSP) problem, which achieves a 15× speedup over an optimized CPU-based implementation. Sridharan et al.~\cite{sridharan2009hardware} present an architecture-efficient solution based on Dijkstra’s algorithm to accelerate the shortest path search, and Takei et al.~\cite{takei2015evaluation} extend this for a high degree of parallelism and large-scale graph search. Recently, Murray et al.~\cite{murray2019programmable} accelerate graph search with the Bellman-Ford algorithm. By leveraging a precomputed roadmap and bounding specific robot quantities, this design enables a more compact and efficient storage structure, dataflows and a low-cost interconnection network.

\section{Partial Reconfiguration}
\label{sec:partial_reconfig}
FPGA technology provides the flexibility of on-site programming and re-programming without going through re-fabrication with a modified design. Partial Reconfiguration (PR) takes this flexibility one step further, allowing the modification of an operating FPGA design by loading a partial configuration file, usually a partial BIT file  \cite{vipin2018fpga}. Using PR, after a full BIT file configures the FPGA, partial BIT files can be downloaded to modify reconfigurable regions in the FPGA without compromising the integrity of the applications running on those parts of the device that are not being reconfigured.

A major performance bottleneck for PR is the configuration overhead, which seriously limits the usefulness of PR. To address this problem, in \cite{liu2010minimizing}, the authors propose a combination of two techniques to minimize the overhead. First, the authors design and implement fully streaming DMA engines to saturate the configuration throughput. Second, the authors exploit a simple form of data redundancy to compress the configuration bitstreams, and implement an intelligent internal configuration access port (ICAP) controller to perform decompression at runtime. This design achieves an effective configuration data transfer throughput of up to 1.2 Gbytes/s, which actually well surpasses the theoretical upper bound of the data transfer throughput, 400 Mbytes/s. Specifically, the proposed fully streaming DMA engines reduce the configuration time from the range of seconds to the range of milliseconds, a more than 1000-fold improvement. In addition, the proposed compression scheme achieves up to a 75\% reduction in bitstream size and results in a decompression circuit with negligible hardware overhead.

Another problem of PR is that it may incur additional energy consumption. In \cite{liu2013achieving},  the authors investigate whether PR can be used to reduce FPGA energy consumption. The core idea is that there are a number of independent circuits within a hardware design, and some can be idle for long periods of time. Idle circuits still consume power though, especially through clock oscillation and static leakage. Using PR, one can replace these circuits during their idle time with others that consume much less power. Since the reconfiguration process itself introduces energy overhead, it is unclear whether this approach actually leads to an overall energy saving or to a loss. This study identifies the precise conditions under which partial reconfiguration reduces the total energy consumption, and proposes solutions to minimize the configuration energy overhead. In this study, PR is compared against clock gating to evaluate its effectiveness. The authors apply these techniques to an existing embedded microprocessor design, and successfully demonstrate that FPGAs can be used to accelerate application performance while also reducing overall energy consumption.

Further, PerceptIn demonstrate in their commercial product that Runtime partial reconfiguration (RPR) is useful for robotic computing, especially computing for autonomous vehicles, because many on-vehicle tasks usually have multiple versions where each is used in a particular scenario \cite{yu2020micro}. For instance, in PerceptIn's design, the localization algorithm relies on salient features; features in key frames are extracted by a feature extraction algorithm (based on ORB features~\cite{rublee2011orb}), whereas features in non-key frames are tracked from previous frames (using optical flow~\cite{lucas1981iterative}); the latter executes in 10 ms, 50\% faster than the former. Spatially sharing the FPGA is not only area-inefficient, but also power-inefficient as the unused portion of the FPGA consumes non-trivial static power. In order to temporally share the FPGA and ``hot-swap'' different algorithms, PerceptIn develop a partial reconfiguration engine (PRE) that dynamically reconfigures part of the FPGA at runtime. The PRE achieves a 400 MB/sec reconfiguration throughput (i.e., bitstream programming rate). Both the feature extraction and tracking bitstreams are less than 4 MB. Thus, the reconfiguration delay is less than 1 ms.
\section{Commercial Applications of FPGAs in Autonomous Vehicles}
\label{sec:case_study1}

\begin{figure*}[t]
\centering
\includegraphics[width=1\columnwidth]{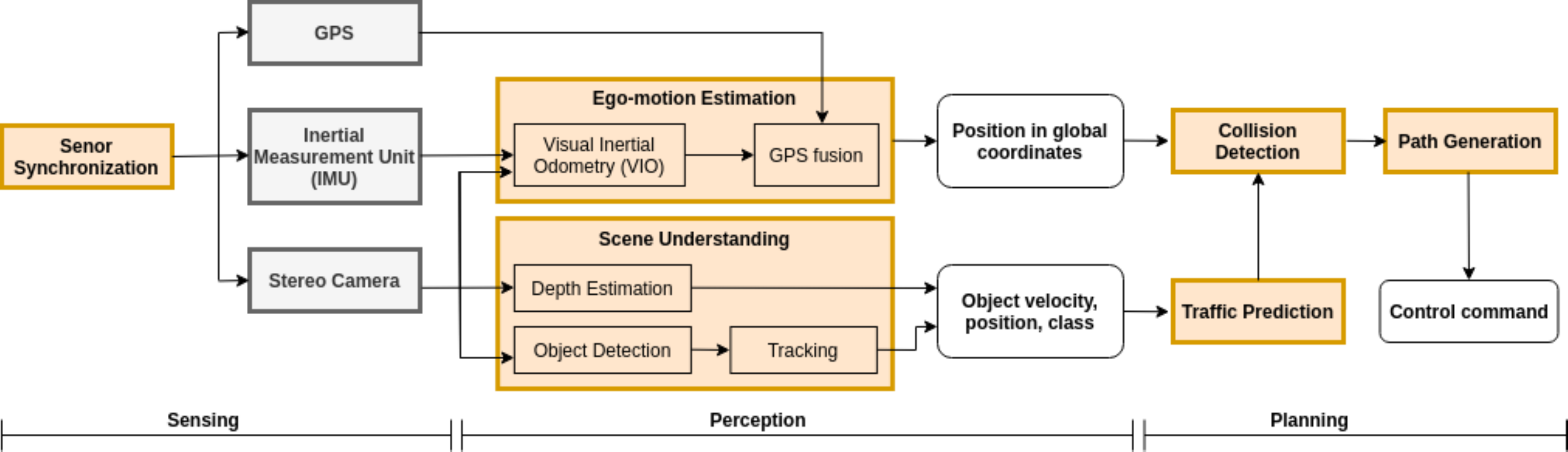}
\caption{Processing pipeline of PerceptIn's on-vehicle processing system.}
\label{fig:pipeline}
\end{figure*}

Over the past three years, PerceptIn has built and commercialized autonomous vehicles for micromobility. Our products have been deployed in China, US, Japan and Switzerland. We summarize system design constraints, workloads and their performance characteristics from the real products. A custom computing system is developed by taking into account the inherent task-level parallisim, cost, safety and programmability\cite{yu2020micro}\cite{fang2018dragonfly+}. FPGA plays a critical role in our system, which synchronizes various sensors and accelerates the component on the critical path. 

\subsection{Computing system}

\textbf{Software pipeline.} Fig. \ref{fig:pipeline} shows the block diagram of the processing pipeline in our vehicle, which consists of three parts: sensing, perception and planning. The sensing module bridges sensors and computing system. It synchronizes various sensor samples for the downstream perception module, which performs two fundamental tasks: 1) locating the vehicle itself in a global map and 2) understanding the surroundings through depth estimation and object detection. The planning module uses the perception results to devise a driveable route, and then converts the planed path into a sequence of control commands, which will drive the vehicle along the path. The control commands are sent to the vehicle's Engine Control Unit (ECU) via the CAN bus interface.

Sensing, perception and planning are serialized. They are all on the critical path of the end-to-end latency. We pipeline the three modules to improve the throughput. Within perception, localization and scene understanding are independent and could execute in parallel. While there are multiple tasks within scene understanding, they are mostly independent with the only exception that object tracking must be serialized with object detection. The task-level parallelisms influence how the tasks are mapped to the hardware platform.

\textbf{Algorithm.}
Our localization module is based on Visual Inertial Odometry algorithms \cite{qin2018vins,sun2018msckf}, which fuses camera images, IMU and GPS samples to estimate the vehicle pose in the global map. The depth estimation employs traditional stereo vision algorithms, which calculates depths according to the principal of triangulation \cite{szeliski2010computer}. In particular, our method is based on the classic ELAS algorithm, which uses hand-crafted features \cite{Elas2010Geiger}. While DNN models for depth estimation exist, they are orders of magnitude more compute-intensive than non-DNN algorithms\cite{feng2019stereo} while providing only marginal accuracy improvements to our use-cases. We detect objects using DNN models, such as YOLO \cite{DBLP:journals/corr/RedmonDGF15}. We use the Kernelized Correlation Filter (KCF) \cite{Hen2015kcf} to track detected objects. The planning algorithm is formulated as Model Predictive Control (MPC)\cite{kelly2013mobile}.

\begin{figure}[t]
\centering
\includegraphics[width=1\columnwidth]{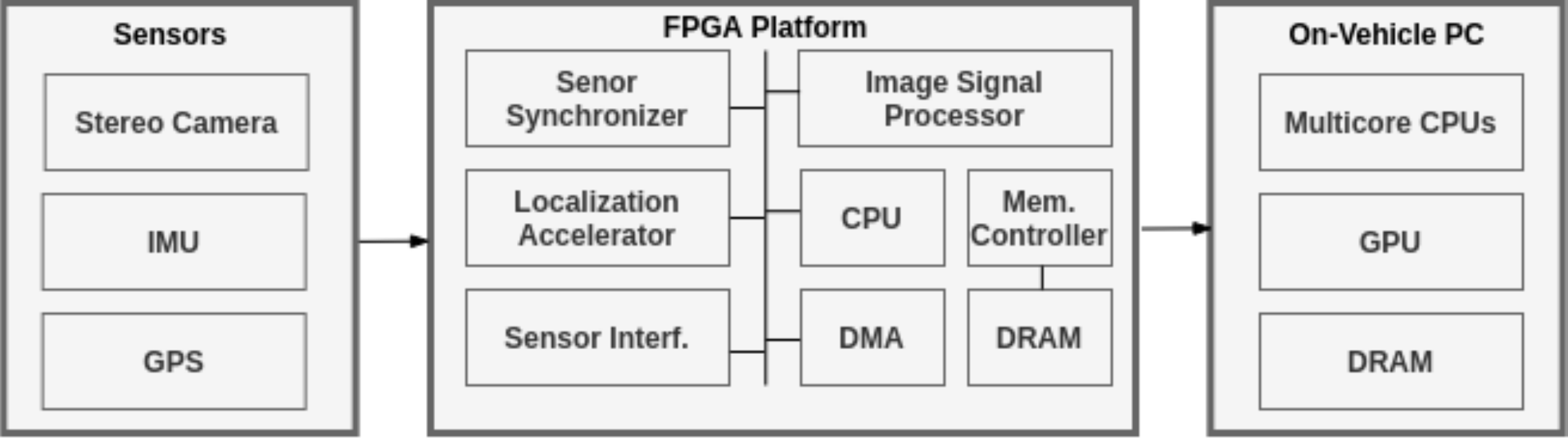}
\caption{The computing system in our autonomous vehicle.}
\label{fig:hwarch}
\end{figure}

\textbf{Hardware architecture.} Fig. \ref{fig:hwarch} is the hardware system designed for our autonomous vehicles. The sensing hardware consists of stereo cameras, IMU and GPS. In particular, our system uses stereo cameras for depth estimation. One of the cameras is also used for semantic tasks such as object detection. The cameras along with the IMU and the GPS drive the VIO-based localization task.

Considering the cost, compute requirements and power budget, our computing platform is composed of a Xilinx Zynq Ultrascale+ FPGA and an on-vehicle PC equipped with an Intel Coffe Lake CPU and an Nvidia GTX 1060 GPU. The PC is the main computing platform, while the FPGA plays a critical role, which bridges sensors and the PC, and provides an acceleration platform. To optimize the end-to-end latency, explore the task level parallelism and ease practical development and deployment, planning and scene understanding are mapped onto the CPU and the GPU respectively, and sensing and localization are implemented on the FPGA platform. 

\subsection{Sensing on FPGA}
We map sensing on the Zynq FPGA platform. The FPGA processes sensor data and transfer sensor data to the PC for subsequent processing. The reason that sensing is mapped to FPGA is three-fold. First, embedded FPGA platforms today are built with rich sensor interface (e.g. standard MIPI Camera Serial Interface) and sensor pre-processing hardware (e.g. ISP). Second, by having the FPGA directly process sensor data in situ, we allow accelerators on the FPGA to directly process sensor data without involving the power-hungry CPU for data movement and task coordination. Finally, processing sensor data on the FPGA naturally leads to a design of hardware-assisted multiple sensor synchronization mechanism.

\textbf{Sensor Synchronization}
Sensor synchronization is critical to perception algorithms that fuse multiple sensors. Sensor fusion algorithms assume sensor samples have been well synchronized. For example, widely adopted datasets, such as KITTI, provide synchronized data so that researchers could focus on algorithmic development.

An ideal synchronization ensures that 1) various sensor samples have a unified timing system, and 2) timestamps of samples precisely record the time of events triggering the sensors. GPS synchronization is now wildly adopted to unify various measurements in a global timing domain. Software-based synchronization associates samples with timestamps at the application or the driver layer. This approach is  inaccurate due to the software processing before the timestamp stage. The software processing introduces variable latency that is non-deterministic. 

To obtain more precise synchronization, we uses a hardware synchronizer implemented by FPGA fabrics. The hardware synchronizer triggers the camera sensors and the IMU using a common timer initialized by the satellite atomic time provided by the GPS device. It records the triggering time of each sensor sample, and then pack the timestamp with the corresponding sensor data. In terms of costs, the synchronizer is extremely lightweight in design with only 1,443 LUTs and 1,587 registers and consumes 5mW of power.

\subsection{Perception on FPGA}
For our autonomous vehicles, the perception tasks includes scene understanding (depth estimation and objection detection) and localization, which are independent. The slower one dictates the overall perception latency.

\begin{figure}[t]
\centering
\includegraphics[width=1\columnwidth]{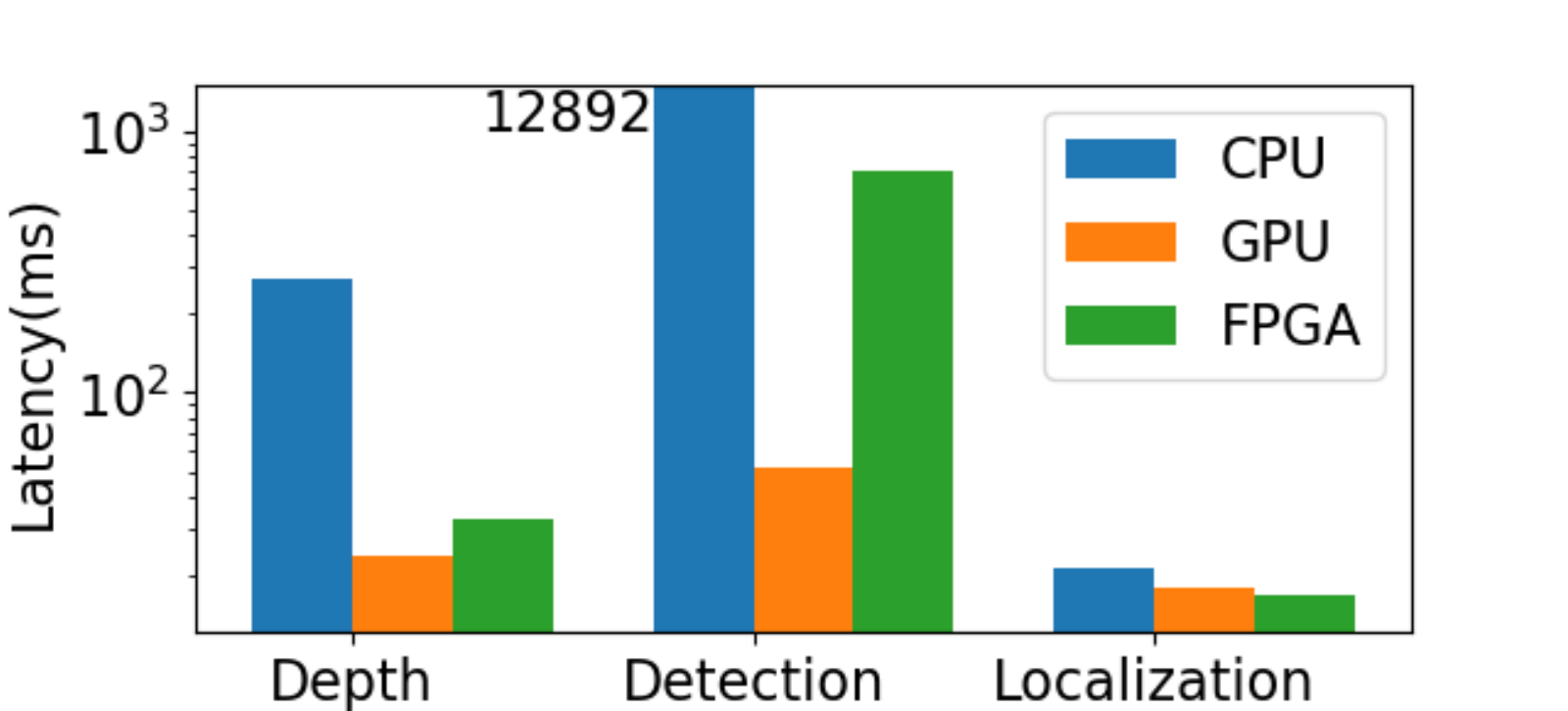}
\caption{Performance comparison of different platforms running three perception tasks.}
\label{fig:perfcomp}
\end{figure}

We evaluate our perception algorithms on the CPU, GPU and Zynq FPGA platform. Fig. \ref{fig:perfcomp} compares the latency of each perception tasks on the FPGA platform with the GPU. Due to the available resources, the FPGA platform is faster than the GPU only for localization, which is more lightweight than other tasks. We offload localization to the FPGA while leaving other perception task on the GPU. This partitioning frees more GPU resources for depth estimation and object detection, which is benefit for reducing the perception pipeline's latency. 

As with classic SLAM algorithms, our localization algorithm consists of a front-end and a back-end. The front-end uses the ORB features and descriptors for detecting and tracking key points \cite{fang2017fpga,tang2018pisoc}. The back-end uses Levenberg-Marquardt's (LM) algorithm, a non-linear optimization algorithm, to optimize the position of 3D key points and the pose of the camera\cite{liu2020pi,qin2019ba}.

The ORB feature extraction/matching and the LM optimizer are the most time-consuming parts of our SLAM algorithm, which take up nearly all the execution time. 
We accelerate ORB feature extraction/matching and the non-linear optimizer on FPGA fabrics. The rest lightweight parts are implemented on the ARM core of the Zyqn platform. We use independent hardware for each camera to extract features and compute descriptors. Hamming distance and Sum of Absoluated Difference (SAD) matching are implemented to obtain stable matching results. Compared with the CPU implementation, our FPGA implementation achieves a 2.2× speedup and 44 fps, and saves 83\% energy.

We use LM algorithm to optimize features and poses over a fixe-size sliding window.
To solve the non-linear optimization problem, the LM algorithm iteratively use Jacobbian to linearize the problem and solve the linear equation at each iteration. Schur elimination is used to reduce the dimension of the linear equation, thus reduce the complexity of solving the equation. Cholesky factorization is employed to solve the linear equation. For sliding-window based vSLAM, the Jacobian and Schur elimination are the most time-consuming parts. By profiling our algorithm on datasets \cite{baldataset}, Schur and Jacobian computations account for 29.8\% and 48.27\% of total time. We implemented Schur elimination and Jacobian updates on FGPA fabrics\cite{liu2020pi}. Compared with the CPU implementation, the FPGA achieves 4× and 27× speedup for Schur and Jacobbian, and saves 76\% energy. 
\section{Application of FPGAs in Space Robotics}
\label{sec:case_study2}

In the 1980s, field-programmable gate arrays (FPGA) emerged as a result of increasing integration in electronics. Before the use of FPGA, glue-logic designs were based on individual boards with fixed components interconnected via a shared standard bus, which has various drawbacks, such as hindrance of high volume data processing and higher susceptibility to radiation-induced errors, in addition to inflexibility. The utilization of FPGAs in space applications began in 1992, for FPGAs offered unprecedented flexibility and significantly reduced the design cycle and development cost \cite{mckerracher1992design}. 

FPGAs can be categorized by the type of their programmable interconnection switches: antifuse, SRAM, and Flash. Each of the three technologies comes with trade-offs. Antifuse FPGAs are non-volatile and have minimal delay due to routing, resulting in a faster speed and lower power consumption. The drawback is evident as they have a relatively more complicated fabrication process and are only one time programmable. SRAM-based FPGAs are the most common type employed in space missions. They are field reprogrammable and use the standard fabrication process that foundries put in significant effort in optimizing, resulting in a faster rate of performance increase. However, based on SRAM, these FPGAs are volatile and may not hold configuration if a power glitch occurs. Also, they have more substantial routing delay, require more power, and have a higher susceptibility to bit errors. Flash-based FPGAs are non-volatile and reprogrammable, and also have low power consumption and route delay. The major drawback is that in-flight reconfiguration is not recommended for flash-based FPGAs due to the potentially destructive results if radiation effects occur during the reconfiguration process \cite{berg2019fpga}. Also, the stability of stored charge on the floating gate is of concern: it is a function including factors such as operating temperature, the electric fields that might disturb the charge. As a result, flash-based FPGAs are not as frequently used in space missions \cite{sheldonflash}.
    
\subsection{Radiation Tolerance for Space Computing}

For electronics intended to operate in space, the harsh space radiation present is an essential factor to consider. Radiation has various effects on electronics, but the commonly focused two are total ionizing dose effect (TID) and single event effects (SEE). TID results from the accumulation of ionizing radiation over time, which causes permanent damage by creating electron-hole pairs in the silicon dioxide layers of MOS devices. The effect of TID is that electronics gradually degrade in their performance parameters and eventually fail to function. Electronics intended for application in space are tested for the total amount of radiation, measured in kRads, they can endure before failure. Usually, electronics that can withstand 100 kRads are sufficient for low earth orbit missions to use for several years \cite{berg2019fpga}.

SEE occurs when high-energy particles from space radiation strike electronics and leave behind an ionized trail. The results are various types of SEEs \cite{gaillard2011single}, which can be categorized as either soft errors, which usually do not cause permanent damage, or hard errors, which often cause permanent damage. Examples of soft error include single event upset (SEU), and single event transient (SET). In SEU, a radiation particle struck a memory element, causing a bit flip. Noteworthy is that as the cell density and clock rate of modern devices increases, multiple cell upset (MCU), corruption of two or more memory cells in a single particle strike, is increasingly becoming a concern. A special type of SEU is single event functional interrupt (SEFI), where the upset leads to loss of normal function of the device by affecting control registers or the clock. In SET, a radiation particle passes through a sensitive node, which generates a transient voltage pulse, causing wrong logic state at the combinatorial logic output. Depending on whether the impact occurs during an active clock edge or not, the error may or may not propagate. Some examples of hard error include single event latch-up (SEL), in which energized particle activates parasitic transistor and then cause a short across the device, and single event burnout (SEB), in which radiation induces high local power dissipation, leading to device failure. In these hard error cases, radiation effects may cause the failure of an entire space mission. 

Space-grade FPGAs can withstand considerable levels of TID and have been designed against most destructive SEEs \cite{wirthlin2013fpgas}. However, SEU susceptibility is pervasive. For the most part, radiation effects on FPGA are not different from those of other CMOS based ICs. The primary anomaly stems from FPGAs’ unique structure, involving programmable interconnections. Depending on their type, FPGAs have different susceptibility toward SEU in their configuration. SRAM FPGAs are designated by NASA as the most susceptible ones due to their volatile nature. Even after the radiation hardening process, the configuration of SRAM FPGAs is only designated as “hardened” or simply having embedded SEE mitigation techniques rather than “hard,”which means close to immune \cite{berg2019fpga}. Configuration SRAM is not used in the same way as the traditional SRAM. A bit flip in configuration causes an instantaneous effect without the need for a read-write cycle. Moreover, instead of producing one single error in the output, the bit flip shifts the user logic directly, changing the device’s behavior. Scrubbing is needed to rectify SRAM configuration. Antifuse and flash FPGAs are less susceptible to effects in configuration and are designated “hard” against SEEs in their configuration without applying radiation hardening techniques \cite{berg2019fpga}. 

Design based SEU/fault mitigation techniques are commonly used, for, in contrast to fabrication level radiation hardening techniques, they can be readily applied to commercial off the shelf (COTS) FPGAs. These techniques can be classified into static and dynamic. Static techniques rely on fault-masking, toleration of error without requiring active fixing. One such example is passive redundancy with voting mechanisms. Dynamic techniques, in contrast, detect faults and act to correct them. The common SEU Mitigation Methods include \cite{brosser2014seu}\cite{ahmed2016fault}:

\begin{enumerate}
	\item Hardware Redundancy: functional blocks are replicated to detect/tolerate faults. Triple modular redundancy (TMR) is perhaps the most widely used mitigation technique. It can be applied to entire processors or parts of circuits. At a circuit level, registers are implemented using three or more flip flops or latches. Then, voters compare the values and output the majority, reducing the likelihood of error due to SEU.  As internal voters are also susceptible to SEU, they are sometimes triplicated also. For mission-critical applications, global signals may be triplicated to mitigate SEUs further. TMR can be implemented at ease with the help supporting HDLs \cite{habinc2002suitability}. It is important to note that a limitation of TMR is that one fault, at most, can be tolerated per voter stage. As a result, TMR is often used with other techniques, such as scrubbing, to prevent error accumulation.
	\item Scrubbing: The vast majority of memory cells in reprogrammable FPGAs contain configuration information. As discussed earlier, configuration memory upset may lead to alteration routing network, loss of function, and other critical effects. Scrubbing, refreshing and restoration of configuration memory to a known-good state, is therefore needed \cite{ahmed2016fault}. The reference configuration memory is usually stored in radiation-hardened memory cells either off or on the device. Scrubbers, processors or configuration controllers, carry out scrubbing. Some advanced SRAM FPGAs, including ones made by Xilinx, support partial reconfiguration, which allows memory repairs to be made without interrupting the operation of the whole device. Scrubbing can be done in frame-level (partial) or device-level (full), which will inevitably lead to some downtime; some devices may not be able to tolerate such an interruption. Blind scrubbing is the most straightforward way of implementation: individual frames are scrubbed periodically without error detection. Blind scrubbing avoids the complexity required in error detection, but extra scrubbing may increase vulnerability to SEUs as errors may be written into frames during the scrubbing process. An alternative to blind scrubbing is readback scrubbing, where scrubbers actively detect errors in configuration through error-correcting code or cyclic redundancy check \cite{brosser2014seu}. If an error is found, scrubber initiates frame-level scrubbing.
\end{enumerate}

Currently, the majority of space-grade FPGA comes from Xilinx and Microsemi. Xilinx offers the Virtex family and Kintex. Both are SRAM based, which have high flexibility. Microsemi offers antifuse based RTAX and Flash-based RTG4, RT PolarFire, which have lower susceptibility against SEE and power consumption. 20 nm Kintex and 28nm RT PolarFire are the latest generations. The European market is offered with Atmel devices and NanoXplore space-grade FPGAs \cite{lentaris2018high}. Table \ref{tab:space-table} shows the specifications of the above devices.

\begin{table*}[t!]
\centering
\renewcommand*{\arraystretch}{1}
\resizebox{\columnwidth}{!}{%
\Huge
\begin{tabular}{c|c|c|c|c|c}
\toprule[1.5pt]
\textbf{Device} & \textbf{Logic} & \textbf{Memory} & \textbf{DSPs} & \textbf{Technology} & \textbf{Rad. Tolerance} \\ \hline
Xilinx Virtex-5QV  & 81.9K LUT6 & 12.3 Mb & 320  & 65 nm SRAM & SEE  immune up to LET$>$100 MeV/(mg$\cdot$cm$^2$) and 1 Mrad TID \\ \hline
Xilinx RT Kintex UltraScale & 331K LUT6  & 38 Mb   & 2760 & 20 nm SRAM      & SEE  immune up to LET$>$80   MeV/(mg$\cdot$cm$^2$) and 100-120 Krads TID \\ \hline
Microsemi RTG4              & 150K LE    & 5 Mb    & 462  & 65 nm Flash     & SEE immune up to LET$>$37 MeV∕(mg$\cdot$cm$^2$) and TID$>$100 Krads        \\ \hline
Microsemi RT PolarFire      & 481K LE    & 33 Mb   & 1480 & 28 nm Flash     & SEE immune up to LET$>$63 MeV∕(mg$\cdot$cm$^2$) and 300 Krads              \\ \hline
Microsemi RTAX              & 4M gates   & 0.5 Mb  & 120  & 150 nm antifuse & SEE immune up to LET$>$37 MeV∕(mg$\cdot$cm$^2$) and 300 Krads TID          \\ \hline
Atmel ATFEE560              & 560K gates & 0.23 Mb & --   & 180 nm SRAM     & SEL immune up to 95 MeV∕(mg$\cdot$cm$^2$)  and 60 Krads TID              \\
\hline
NanoXplore NG-LARGE         & 137K LUT4  & 9.2 Mb  & 384  & 65 nm SRAM      & SEL immune up to 60 MeV∕(mg$\cdot$cm$^2$)  and 100 Krads TID    \\  
\bottomrule[1.5pt]
\end{tabular}%
\caption{Specifications of Space-Grade FPGAs.}
\label{tab:space-table}
}
\end{table*}

\subsection{FPGAs in Space Missions}

For space robotics, processing power is of particular importance, given the range of information required to accurately and efficiently process. Many of the current and previous space missions are packed with sophisticated algorithms that are mostly static. They serve to increase the efficiency of data transmission; nevertheless, data processing is done mainly on the ground. As the travel distance of missions increases, transmitting all data to, and processing it on the ground is no longer an efficient or even viable option due to transmission delay. As a result, space robots need to become more adaptable and autonomous. They will also need to pre-process on-board a large amount of data collected and compress it before sending it back to Earth \cite{li2019enabling}.

The rapid development of new generation FPGAs may fill the need in space robotics. FPGAs enable robotic systems to be reconfigurable in real-time, making the systems more adaptable by allowing them to respond more efficiently to changes in environment and data. As a result, autonomous reconfiguration and performance optimization can be achieved. Also, the FPGAs have a high capability for parallel processing, which is useful in boosting processing performance. The use of FPGA is present in various space robots. Some of the most prominent examples of the application are the NASA Mars rovers. Since the first pair of rovers were launched in 2003, the presence of FPGAs have steadily increased in the later rovers.

\subsubsection{Mars Exploration Rover Missions}

Beginning in the early 2000s, NASA have been using FPGAs in exploration rover control and lander control. In Opportunity and Spirit, the two Mars rovers launched in 2003, two Xilinx Virtex XQVR1000s were in the motor control board \cite{ratter2004fpgas}, which operates motors on instruments as well as rover wheels. In addition, an Actel RT 1280 FPGA was used in each of the 20 cameras on the rovers to receive and dispatch hardware commands. The camera electronics consist of clock driver that provides timing pulses through the charge-coupled device (CCD), an IC containing an array of linked or coupled capacitors. Also, there are signal chains that amplify the CCD output and convert it from analog to digital. The Actel FPGA provides the timing, logic, and control functions in the CCD signal chain and inserts a camera ID into camera telemetry to simplify processing \cite{bell2003mars}. 

Selected electronic parts have to undergo a multi-step flight consideration process before utilized in any space exploration mission \cite{ratter2004fpgas}\cite{sfsdet}. The first step is the general flight approval, during which the manufacturers perform additional space-grade verification tests beyond the normal commercial evaluation, and NASA meticulously examines the results. Additional device parameters, such as temperature considerations and semiconductor characteristics are verified in these tests. What follows is flight-specific approval. In this step, NASA engineers examine the device compatibility with the mission. For instance, considerations of the operating environment including factors like temperature and radiation. Also included are a variety of mission-specific situations that the robot may encounter and the associated risk assessment. Depending on the specific application of the device, whether mission critical or not, and the expected mission lifetime, the risk standards varies. Finally, parts go through specific design consideration to ensure all the design requirements have been met. Parts are examined for their designs addressing issues such as SEL, SEU, SEFI. The Xilinx FPGAs used addressed some of the SEE through the following methods \cite{bell2003mars}:
\begin{enumerate}
	\item Fabrication processes largely prevents SEL
	\item TMR reduces SEU frequency 
	\item Scrubbing allows device recovery from single event functional interrupts
\end{enumerate}

MER went successful and despite being designed for only 90 Martian days (1 Martian day = 24.6 hours), continued until 2019. The implementation of mitigation techniques was also proven to be effective as the observed error rate was very similar to that predicted \cite{ratter2004fpgas}. 

\subsubsection{Mars Science Laboratory Mission}
Launched in 2011, Mars Science Lab (MSL) was the new Rover sent on to Mars. FPGAs were heavily used in its key components, mainly responsible for scientific instrument control, image processing, and communications. 

Curiosity has 17 cameras on board: four navigation cameras, eight hazard cameras, the Mars Hand Lens Imager (MAHLI), two Mast Cameras, the Mars Descent Imager (MARDI), and the ChemCam Remote Microscopic Imager \cite{malin2017mars}. MAHLI, the mast cameras, and MARDI share the same electronics design. Similar to the system used on MER, an Actel FPGA provides the timing, logic, and control functions in the CCD signal chain and transmits pixels to the digital electronics assembly (DEA), which interfaces the camera heads with the rover electronics, transmitting command to the camera heads and data back to the rover. There is one DEA dedicated to each of the imagers above. Each is has a Virtex-II FPGA that contains a Microblaze soft-processor core. All of the core functionalities of the DEA, including timing, interface, and compression, are implemented in the FPGA as logic peripherals of the Microblaze. Specifically, the DEA provides an image processing pipeline that includes 12 to 8-bit commanding of input pixels, horizontal subframing, and lossless or JPEG image compression \cite{malin2017mars}. What runs on the Microblaze is the DEA flight software, which coordinates DEA hardware functions such as camera movements. It receives and executes commands, and transmits command from the Earth. The flight software also implements image acquisition algorithms, including autofocus and autoexposure, performs error correction of flash memory, and mechanism control fault protection \cite{malin2017mars}. In total, the flight software consists of 10,000 lines of ANSI C code, all implemented on the FPGA. Additionally, FPGAs power communication boxes (Electra-Lite) to provide critical communication to Earth from the rovers through a Mars relay network \cite{edwards2003electra}. They are responsible for a variety of high speed bulk signal processing. 

\subsubsection{Mars 2020 Mission}

Perseverance is NASA’s latest launched Mars rover. The presence of FPGA continued and increased. FPGA was used in the autonomous driving system as a coprocessor for algorithm acceleration for the first time in NASA’s planetary rovers. Perseverance runs on the GESTALT (grid-based estimation of surface traversability applied to local terrain) AutoNav algorithm same as Curiosity \cite{johnson2017lander}. Added was the FPGA based accelerator, called Vision Compute Element (VCE). During landing, VCE serves to provide sufficient computing power for the Lander Vision System (LVS), which performs an intensive task of estimates the landing location in 10 seconds by fusing data from the designed landing location, IMU, and landmark matches. After landing, the connection between VCE and LVS is severed. Instead, VCE is repurposed for the GESTALT driving algorithm. The VCE has three cards plugged into a PCI backplane: a CPU card with BAE RAD750 processor, a Compute Element Power Conditioning Unit (CEPCU), and a Computer Vision Acceleration Card (CVAC). While the former two parts were inherited from the MLS mission, the CVAC is new. It has two FPGAs. One is called the Vision Processor--a Xilinx Virtex 5QV that contains image processing modules for matching landmarks to estimate position. The other is called the Housekeeping FPGA--a Microsemi RTAX 2000 antifuse FPGA that handles tasks such as synchronization with the spacecraft, power management, Vision Processor configuration. 

Through more than two decades of use in space, FPGAs have shown their reliability and applicability for space robotic missions. The properties of FPGAs make them good onboard processors, ones that have high reliability, adaptability, processing power, and power efficiency: FPGAs have been used for space robotic missions for decades and are proven in reliability; they have unrivaled adaptability and can even be reconfigured in run time; their capability for high degree parallel processing allow significant acceleration in executing many complex algorithms; hardware/software co-design method makes them potentially more power-efficient. They may finally help us close the two-decade performance gap between commercial processors and space-grade ASICs. As a direct result, the achievements that the world has made in fields such as deep learning and computer vision, which were often too computationally intense for space-grade processors to be used, may become applicable for robots in space in the near future. The implementation of those new technologies will be of great benefit for space robots, boosting their autonomy and capabilities and allowing us to explore farther and faster.
\section{Conclusion}
\label{sec:conclusion}
In this paper, we review the state-of-the-art FPGA-based robotic computing accelerator designs and summarize their adopted optimized techniques. According to the results shown in Section~\ref{sec:perception}, \ref{sec:localization} and~\ref{sec:planning_control}, by co-designing both the software and hardware, FPGA can achieve more than 10× better performance and energy efficiency compared to the CPU and GPU implementations. We also review the partial reconfiguration methodology in FPGA implementation to further improve the design flexibility and reduce the overhead.  Finally, by presenting some recent FPGA-based robotics applications in commercial and space areas, we demonstrate that FPGA has excellent potential and is a promising candidate for robotic computing acceleration due to its high reliability, adaptability and power efficiency.

The authors believe that FPGAs are the best compute substrate for robotic applications for several reasons: first, robotic algorithms are still evolving rapidly, and thus any ASIC-based accelerators will be months or even years behind the state-of-the-art algorithms; on the other hand, FPGAs can be dynamically updated as needed.  Second, robotic workloads are highly diverse, thus it is difficult for any ASIC-based robotic computing accelerator to reach economies of scale in the near future; on the other hand, FPGAs are a cost-effective and energy-effective alternative before one type of accelerator reaches economies of scale. Third, compared to SoCs that have reached economies of scale, e.g., mobile SoCs, FPGAs deliver a significant performance advantage. Fourth, partial reconfiguration allows multiple robotic workloads to time-share an FPGA, thus allowing one chip to serve multiple applications, leading to overall cost and energy reduction.

However, FPGAs are still not the mainstream computing substrate for robotic workloads, for several reasons: first, FPGA programming is still much more challenging than regular software programming, and the supply of FPGA engineers is still limited. Second, although there is significant progress in the past few years in the FPGA High-Level Synthesis (HLS) automation, such as \cite{vivado}, HLS is still not able to produce optimized code, and IP supports for robotic workloads are still extremely limited.  Third, commercial software support for robotic workloads on FPGAs is still missing. For instance, there is no official ROS support on any commercial FPGA platform today. For robotic companies to fully exploit the power of FPGAs, these problems need to be first addressed, and the authors use these problems to motivate our future research work.

\ifCLASSOPTIONcaptionsoff
  \newpage
\fi



%


\bibliographystyle{ieeetr}
\bibliography{refs}

\begin{thebibliography}{100}

\bibitem{qiantori2012emergency}
A.~Qiantori, A.~B. Sutiono, H.~Hariyanto, H.~Suwa, and T.~Ohta, ``An emergency
  medical communications system by low altitude platform at the early stages of
  a natural disaster in indonesia,'' {\em Journal of medical systems}, vol.~36,
  no.~1, pp.~41--52, 2012.

\bibitem{ryan2005mode}
A.~Ryan and J.~K. Hedrick, ``A mode-switching path planner for uav-assisted
  search and rescue,'' in {\em Proceedings of the 44th IEEE Conference on
  Decision and Control}, pp.~1471--1476, IEEE, 2005.

\bibitem{smolyanskiy2017toward}
N.~Smolyanskiy, A.~Kamenev, J.~Smith, and S.~Birchfield, ``Toward low-flying
  autonomous mav trail navigation using deep neural networks for environmental
  awareness,'' in {\em 2017 IEEE/RSJ International Conference on Intelligent
  Robots and Systems (IROS)}, pp.~4241--4247, IEEE, 2017.

\bibitem{giusti2015machine}
A.~Giusti, J.~Guzzi, D.~C. Cire{\c{s}}an, F.-L. He, J.~P. Rodr{\'\i}guez,
  F.~Fontana, M.~Faessler, C.~Forster, J.~Schmidhuber, G.~Di~Caro, {\em
  et~al.}, ``A machine learning approach to visual perception of forest trails
  for mobile robots,'' {\em IEEE Robotics and Automation Letters}, vol.~1,
  no.~2, pp.~661--667, 2015.

\bibitem{stolaroff2018energy}
J.~K. Stolaroff, C.~Samaras, E.~R. O’Neill, A.~Lubers, A.~S. Mitchell, and
  D.~Ceperley, ``Energy use and life cycle greenhouse gas emissions of drones
  for commercial package delivery,'' {\em Nature communications}, vol.~9,
  no.~1, pp.~1--13, 2018.

\bibitem{kim2018survey}
S.~J. Kim, Y.~Jeong, S.~Park, K.~Ryu, and G.~Oh, ``A survey of drone use for
  entertainment and avr (augmented and virtual reality),'' in {\em Augmented
  Reality and Virtual Reality}, pp.~339--352, Springer, 2018.

\bibitem{jung2018direct}
S.~Jung, S.~Cho, D.~Lee, H.~Lee, and D.~H. Shim, ``A direct visual
  servoing-based framework for the 2016 iros autonomous drone racing
  challenge,'' {\em Journal of Field Robotics}, vol.~35, no.~1, pp.~146--166,
  2018.

\bibitem{FAA2020}
``Fact sheet – the federal aviation administration (faa) aerospace forecast
  fiscal years (fy) 2020-2040.''
  \url{https://www.faa.gov/news/fact_sheets/news_story.cfm?newsId=24756}, 2020.

\bibitem{liu2017creating}
S.~Liu, L.~Li, J.~Tang, S.~Wu, and J.-L. Gaudiot, ``Creating autonomous vehicle
  systems,'' {\em Synthesis Lectures on Computer Science}, vol.~6, no.~1,
  pp.~i--186, 2017.

\bibitem{krishnan2020sky}
S.~Krishnan, Z.~Wan, K.~Bhardwaj, P.~Whatmough, A.~Faust, G.-Y. Wei, D.~Brooks,
  and V.~J. Reddi, ``The sky is not the limit: A visual performance model for
  cyber-physical co-design in autonomous machines,'' {\em IEEE Computer
  Architecture Letters}, vol.~19, no.~1, pp.~38--42, 2020.

\bibitem{krishnan2021machine}
S.~Krishnan, Z.~Wan, K.~Bharadwaj, P.~Whatmough, A.~Faust, S.~Neuman, G.-Y.
  Wei, D.~Brooks, and V.~J. Reddi, ``Machine learning-based automated design
  space exploration for autonomous aerial robots,'' {\em arXiv preprint
  arXiv:2102.02988}, 2021.

\bibitem{liu2020autonomous}
S.~Liu and J.-L. Gaudiot, ``Autonomous vehicles lite self-driving technologies
  should start small, go slow,'' {\em IEEE Spectrum}, vol.~57, no.~3,
  pp.~36--49, 2020.

\bibitem{liu2019edge}
S.~Liu, L.~Liu, J.~Tang, B.~Yu, Y.~Wang, and W.~Shi, ``Edge computing for
  autonomous driving: Opportunities and challenges,'' {\em Proceedings of the
  IEEE}, vol.~107, no.~8, pp.~1697--1716, 2019.

\bibitem{liu2017computer}
S.~Liu, J.~Tang, Z.~Zhang, and J.-L. Gaudiot, ``Computer architectures for
  autonomous driving,'' {\em Computer}, vol.~50, no.~8, pp.~18--25, 2017.

\bibitem{guo2019dl}
K.~Guo, S.~Zeng, J.~Yu, Y.~Wang, and H.~Yang, ``[dl] a survey of fpga-based
  neural network inference accelerators,'' {\em ACM Transactions on
  Reconfigurable Technology and Systems (TRETS)}, vol.~12, no.~1, pp.~1--26,
  2019.

\bibitem{yu2020micro}
B.~Yu, W.~Hu, L.~Xu, J.~Tang, S.~Liu, and Y.~Zhu, ``Building the computing
  system for autonomous micromobility vehicles: Designconstraints and
  architectural optimizations,'' in {\em 2020 53rd Annual IEEE/ACM
  International Symposium on Microarchitecture (MICRO)}, IEEE, 2020.

\bibitem{1467360}
N.~{Dalal} and B.~{Triggs}, ``Histograms of oriented gradients for human
  detection,'' in {\em 2005 IEEE Computer Society Conference on Computer Vision
  and Pattern Recognition (CVPR'05)}, vol.~1, pp.~886--893 vol. 1, 2005.

\bibitem{1315232}
{Xuming He}, R.~S. {Zemel}, and M.~A. {Carreira-Perpinan}, ``Multiscale
  conditional random fields for image labeling,'' in {\em Proceedings of the
  2004 IEEE Computer Society Conference on Computer Vision and Pattern
  Recognition, 2004. CVPR 2004.}, vol.~2, pp.~II--II, 2004.

\bibitem{101007}
X.~He, R.~S. Zemel, and D.~Ray, ``Learning and incorporating top-down cues in
  image segmentation,'' in {\em Computer Vision -- ECCV 2006} (A.~Leonardis,
  H.~Bischof, and A.~Pinz, eds.), (Berlin, Heidelberg), pp.~338--351, Springer
  Berlin Heidelberg, 2006.

\bibitem{7410891}
Y.~{Xiang}, A.~{Alahi}, and S.~{Savarese}, ``Learning to track: Online
  multi-object tracking by decision making,'' in {\em 2015 IEEE International
  Conference on Computer Vision (ICCV)}, pp.~4705--4713, 2015.

\bibitem{Girshick_2015}
R.~Girshick, ``Fast r-cnn,'' {\em 2015 IEEE International Conference on
  Computer Vision (ICCV)}, Dec 2015.

\bibitem{DBLP:journals/corr/RenHG015}
S.~Ren, K.~He, R.~B. Girshick, and J.~Sun, ``Faster {R-CNN:} towards real-time
  object detection with region proposal networks,'' {\em CoRR},
  vol.~abs/1506.01497, 2015.

\bibitem{DBLP:journals/corr/LiuAESR15}
W.~Liu, D.~Anguelov, D.~Erhan, C.~Szegedy, S.~E. Reed, C.~Fu, and A.~C. Berg,
  ``{SSD:} single shot multibox detector,'' {\em CoRR}, vol.~abs/1512.02325,
  2015.

\bibitem{DBLP:journals/corr/RedmonDGF15}
J.~Redmon, S.~K. Divvala, R.~B. Girshick, and A.~Farhadi, ``You only look once:
  Unified, real-time object detection,'' {\em CoRR}, vol.~abs/1506.02640, 2015.

\bibitem{DBLP:journals/corr/RedmonF16}
J.~Redmon and A.~Farhadi, ``{YOLO9000:} better, faster, stronger,'' {\em CoRR},
  vol.~abs/1612.08242, 2016.

\bibitem{DBLP:journals/corr/LongSD14}
J.~Long, E.~Shelhamer, and T.~Darrell, ``Fully convolutional networks for
  semantic segmentation,'' {\em CoRR}, vol.~abs/1411.4038, 2014.

\bibitem{DBLP:journals/corr/HeZR014}
K.~He, X.~Zhang, S.~Ren, and J.~Sun, ``Spatial pyramid pooling in deep
  convolutional networks for visual recognition,'' {\em CoRR},
  vol.~abs/1406.4729, 2014.

\bibitem{DBLP:journals/corr/ZhaoSQWJ16}
H.~Zhao, J.~Shi, X.~Qi, X.~Wang, and J.~Jia, ``Pyramid scene parsing network,''
  {\em CoRR}, vol.~abs/1612.01105, 2016.

\bibitem{DBLP:journals/corr/BertinettoVHVT16}
L.~Bertinetto, J.~Valmadre, J.~F. Henriques, A.~Vedaldi, and P.~H.~S. Torr,
  ``Fully-convolutional siamese networks for object tracking,'' {\em CoRR},
  vol.~abs/1606.09549, 2016.

\bibitem{1638022}
H.~{Durrant-Whyte} and T.~{Bailey}, ``Simultaneous localization and mapping:
  part i,'' {\em IEEE Robotics Automation Magazine}, vol.~13, no.~2,
  pp.~99--110, 2006.

\bibitem{montemerlo2008junior}
M.~Montemerlo, J.~Becker, S.~Bhat, H.~Dahlkamp, D.~Dolgov, S.~Ettinger,
  D.~Haehnel, T.~Hilden, G.~Hoffmann, B.~Huhnke, {\em et~al.}, ``Junior: The
  stanford entry in the urban challenge,'' {\em Journal of field Robotics},
  vol.~25, no.~9, pp.~569--597, 2008.

\bibitem{6803933}
J.~{Ziegler}, P.~{Bender}, M.~{Schreiber}, H.~{Lategahn}, T.~{Strauss},
  C.~{Stiller}, T.~{Dang}, U.~{Franke}, N.~{Appenrodt}, C.~G. {Keller},
  E.~{Kaus}, R.~G. {Herrtwich}, C.~{Rabe}, D.~{Pfeiffer}, F.~{Lindner},
  F.~{Stein}, F.~{Erbs}, M.~{Enzweiler}, C.~{Knöppel}, J.~{Hipp}, M.~{Haueis},
  M.~{Trepte}, C.~{Brenk}, A.~{Tamke}, M.~{Ghanaat}, M.~{Braun}, A.~{Joos},
  H.~{Fritz}, H.~{Mock}, M.~{Hein}, and E.~{Zeeb}, ``Making bertha drive—an
  autonomous journey on a historic route,'' {\em IEEE Intelligent
  Transportation Systems Magazine}, vol.~6, no.~2, pp.~8--20, 2014.

\bibitem{katrakazas2015real}
C.~Katrakazas, M.~Quddus, W.-H. Chen, and L.~Deka, ``Real-time motion planning
  methods for autonomous on-road driving: State-of-the-art and future research
  directions,'' {\em Transportation Research Part C: Emerging Technologies},
  vol.~60, pp.~416--442, 2015.

\bibitem{paden2016survey}
B.~Paden, M.~{\v{C}}{\'a}p, S.~Z. Yong, D.~Yershov, and E.~Frazzoli, ``A survey
  of motion planning and control techniques for self-driving urban vehicles,''
  {\em IEEE Transactions on intelligent vehicles}, vol.~1, no.~1, pp.~33--55,
  2016.

\bibitem{deng2012fuzzy}
Y.~Deng, Y.~Chen, Y.~Zhang, and S.~Mahadevan, ``Fuzzy dijkstra algorithm for
  shortest path problem under uncertain environment,'' {\em Applied Soft
  Computing}, vol.~12, no.~3, pp.~1231--1237, 2012.

\bibitem{hart1968formal}
P.~E. Hart, N.~J. Nilsson, and B.~Raphael, ``A formal basis for the heuristic
  determination of minimum cost paths,'' {\em IEEE transactions on Systems
  Science and Cybernetics}, vol.~4, no.~2, pp.~100--107, 1968.

\bibitem{lavalle2001randomized}
S.~M. LaValle and J.~J. Kuffner~Jr, ``Randomized kinodynamic planning,'' {\em
  The international journal of robotics research}, vol.~20, no.~5,
  pp.~378--400, 2001.

\bibitem{kavraki1996probabilistic}
L.~E. Kavraki, P.~Svestka, J.-C. Latombe, and M.~H. Overmars, ``Probabilistic
  roadmaps for path planning in high-dimensional configuration spaces,'' {\em
  IEEE transactions on Robotics and Automation}, vol.~12, no.~4, pp.~566--580,
  1996.

\bibitem{shalev2016long}
S.~Shalev-Shwartz, N.~Ben-Zrihem, A.~Cohen, and A.~Shashua, ``Long-term
  planning by short-term prediction,'' {\em arXiv preprint arXiv:1602.01580},
  2016.

\bibitem{gomez2012optimal}
M.~G{\'o}mez, R.~Gonz{\'a}lez, T.~Mart{\'\i}nez-Mar{\'\i}n, D.~Meziat, and
  S.~S{\'a}nchez, ``Optimal motion planning by reinforcement learning in
  autonomous mobile vehicles,'' {\em Robotica}, vol.~30, no.~2, p.~159, 2012.

\bibitem{shalev2016safe}
S.~Shalev-Shwartz, S.~Shammah, and A.~Shashua, ``Safe, multi-agent,
  reinforcement learning for autonomous driving,'' {\em arXiv preprint
  arXiv:1610.03295}, 2016.

\bibitem{bojarski2016end}
M.~Bojarski, D.~Del~Testa, D.~Dworakowski, B.~Firner, B.~Flepp, P.~Goyal, L.~D.
  Jackel, M.~Monfort, U.~Muller, J.~Zhang, {\em et~al.}, ``End to end learning
  for self-driving cars,'' {\em arXiv preprint arXiv:1604.07316}, 2016.

\bibitem{geng2017scenario}
X.~Geng, H.~Liang, B.~Yu, P.~Zhao, L.~He, and R.~Huang, ``A scenario-adaptive
  driving behavior prediction approach to urban autonomous driving,'' {\em
  Applied Sciences}, vol.~7, no.~4, p.~426, 2017.

\bibitem{watkins1992q}
C.~J. Watkins and P.~Dayan, ``Q-learning,'' {\em Machine learning}, vol.~8,
  no.~3-4, pp.~279--292, 1992.

\bibitem{konda2000actor}
V.~R. Konda and J.~N. Tsitsiklis, ``Actor-critic algorithms,'' in {\em Advances
  in neural information processing systems}, pp.~1008--1014, 2000.

\bibitem{hicks2013depth}
S.~L. Hicks, I.~Wilson, L.~Muhammed, J.~Worsfold, S.~M. Downes, and C.~Kennard,
  ``A depth-based head-mounted visual display to aid navigation in partially
  sighted individuals,'' {\em PloS one}, vol.~8, no.~7, p.~e67695, 2013.

\bibitem{whelan2016elasticfusion}
T.~Whelan, R.~F. Salas-Moreno, B.~Glocker, A.~J. Davison, and S.~Leutenegger,
  ``Elasticfusion: Real-time dense slam and light source estimation,'' {\em The
  International Journal of Robotics Research}, vol.~35, no.~14, pp.~1697--1716,
  2016.

\bibitem{prisacariu2017infinitam}
V.~A. Prisacariu, O.~K{\"a}hler, S.~Golodetz, M.~Sapienza, T.~Cavallari, P.~H.
  Torr, and D.~W. Murray, ``Infinitam v3: A framework for large-scale 3d
  reconstruction with loop closure,'' {\em arXiv preprint arXiv:1708.00783},
  2017.

\bibitem{golodetz2018collaborative}
S.~Golodetz, T.~Cavallari, N.~A. Lord, V.~A. Prisacariu, D.~W. Murray, and
  P.~H. Torr, ``Collaborative large-scale dense 3d reconstruction with online
  inter-agent pose optimisation,'' {\em IEEE transactions on visualization and
  computer graphics}, vol.~24, no.~11, pp.~2895--2905, 2018.

\bibitem{perez2019fpga}
M.~P{\'e}rez-Patricio and A.~Aguilar-Gonz{\'a}lez, ``Fpga implementation of an
  efficient similarity-based adaptive window algorithm for real-time stereo
  matching,'' {\em Journal of Real-Time Image Processing}, vol.~16, no.~2,
  pp.~271--287, 2019.

\bibitem{yang2014depth}
D.-W. Yang, L.-C. Chu, C.-W. Chen, J.~Wang, and M.-D. Shieh,
  ``Depth-reliability-based stereo-matching algorithm and its vlsi architecture
  design,'' {\em IEEE Transactions on Circuits and Systems for Video
  Technology}, vol.~25, no.~6, pp.~1038--1050, 2014.

\bibitem{aguilar2016fpga}
A.~Aguilar-Gonz{\'a}lez and M.~Arias-Estrada, ``An fpga stereo matching
  processor based on the sum of hamming distances,'' in {\em International
  symposium on applied reconfigurable computing}, pp.~66--77, Springer, 2016.

\bibitem{perez2016fpga}
M.~P{\'e}rez-Patricio, A.~Aguilar-Gonz{\'a}lez, M.~Arias-Estrada, H.-R.
  Hernandez-de Leon, J.-L. Camas-Anzueto, and J.~de~Jes{\'u}s
  Osuna-Couti{\~n}o, ``An fpga stereo matching unit based on fuzzy logic,''
  {\em Microprocessors and Microsystems}, vol.~42, pp.~87--99, 2016.

\bibitem{cocorullo2016efficient}
G.~Cocorullo, P.~Corsonello, F.~Frustaci, and S.~Perri, ``An efficient
  hardware-oriented stereo matching algorithm,'' {\em Microprocessors and
  Microsystems}, vol.~46, pp.~21--33, 2016.

\bibitem{santos2016scalable}
P.~M. Santos, J.~C. Ferreira, and J.~S. Matos, ``Scalable hardware architecture
  for disparity map computation and object location in real-time,'' {\em
  Journal of Real-Time Image Processing}, vol.~11, no.~3, pp.~473--485, 2016.

\bibitem{ali2017exploring}
K.~M. Ali, R.~B. Atitallah, N.~Fakhfakh, and J.-L. Dekeyser, ``Exploring hls
  optimizations for efficient stereo matching hardware implementation,'' in
  {\em International Symposium on Applied Reconfigurable Computing},
  pp.~168--176, Springer, 2017.

\bibitem{mccullagh2012real}
B.~McCullagh, ``Real-time disparity map computation using the cell broadband
  engine,'' {\em Journal of Real-Time Image Processing}, vol.~7, no.~2,
  pp.~87--93, 2012.

\bibitem{li20173d}
L.~Li, X.~Yu, S.~Zhang, X.~Zhao, and L.~Zhang, ``3d cost aggregation with
  multiple minimum spanning trees for stereo matching,'' {\em Applied optics},
  vol.~56, no.~12, pp.~3411--3420, 2017.

\bibitem{zha2016real}
D.~Zha, X.~Jin, and T.~Xiang, ``A real-time global stereo-matching on fpga,''
  {\em Microprocessors and Microsystems}, vol.~47, pp.~419--428, 2016.

\bibitem{puglia2017real}
L.~Puglia, M.~Vigliar, and G.~Raiconi, ``Real-time low-power fpga architecture
  for stereo vision,'' {\em IEEE Transactions on Circuits and Systems II:
  Express Briefs}, vol.~64, no.~11, pp.~1307--1311, 2017.

\bibitem{kjaer2010two}
A.~Kj{\ae}r-Nielsen, K.~Pauwels, J.~B. Jessen, M.~Van~Hulle, N.~Kr{\"u}ger,
  {\em et~al.}, ``A two-level real-time vision machine combining coarse-and
  fine-grained parallelism,'' {\em Journal of Real-Time Image Processing},
  vol.~5, no.~4, pp.~291--304, 2010.

\bibitem{wong2002sum}
S.~Wong, S.~Vassiliadis, and S.~Cotofana, ``A sum of absolute differences
  implementation in fpga hardware,'' in {\em Proceedings. 28th Euromicro
  Conference}, pp.~183--188, IEEE, 2002.

\bibitem{hisham2015template}
M.~Hisham, S.~N. Yaakob, R.~A. Raof, A.~A. Nazren, and N.~W. Embedded,
  ``Template matching using sum of squared difference and normalized cross
  correlation,'' in {\em 2015 IEEE Student Conference on Research and
  Development (SCOReD)}, pp.~100--104, IEEE, 2015.

\bibitem{yoo2009fast}
J.-C. Yoo and T.~H. Han, ``Fast normalized cross-correlation,'' {\em Circuits,
  systems and signal processing}, vol.~28, no.~6, p.~819, 2009.

\bibitem{froba2004face}
B.~Froba and A.~Ernst, ``Face detection with the modified census transform,''
  in {\em Sixth IEEE International Conference on Automatic Face and Gesture
  Recognition, 2004. Proceedings.}, pp.~91--96, IEEE, 2004.

\bibitem{jin2009fpga}
S.~Jin, J.~Cho, X.~Dai~Pham, K.~M. Lee, S.-K. Park, M.~Kim, and J.~W. Jeon,
  ``Fpga design and implementation of a real-time stereo vision system,'' {\em
  IEEE transactions on circuits and systems for video technology}, vol.~20,
  no.~1, pp.~15--26, 2009.

\bibitem{zhang2011real}
L.~Zhang, K.~Zhang, T.~S. Chang, G.~Lafruit, G.~K. Kuzmanov, and D.~Verkest,
  ``Real-time high-definition stereo matching on fpga,'' in {\em Proceedings of
  the 19th ACM/SIGDA international symposium on Field programmable gate
  arrays}, pp.~55--64, 2011.

\bibitem{honegger2012real}
D.~Honegger, P.~Greisen, L.~Meier, P.~Tanskanen, and M.~Pollefeys, ``Real-time
  velocity estimation based on optical flow and disparity matching,'' in {\em
  2012 IEEE/RSJ International Conference on Intelligent Robots and Systems},
  pp.~5177--5182, IEEE, 2012.

\bibitem{jin2014fast}
M.~Jin and T.~Maruyama, ``Fast and accurate stereo vision system on fpga,''
  {\em ACM Transactions on Reconfigurable Technology and Systems (TRETS)},
  vol.~7, no.~1, pp.~1--24, 2014.

\bibitem{park2007real}
S.~Park and H.~Jeong, ``Real-time stereo vision fpga chip with low error
  rate,'' in {\em 2007 International Conference on Multimedia and Ubiquitous
  Engineering (MUE'07)}, pp.~751--756, IEEE, 2007.

\bibitem{sabihuddin2008dynamic}
S.~Sabihuddin, J.~Islam, and W.~J. MacLean, ``Dynamic programming approach to
  high frame-rate stereo correspondence: A pipelined architecture implemented
  on a field programmable gate array,'' in {\em 2008 Canadian Conference on
  Electrical and Computer Engineering}, pp.~001461--001466, IEEE, 2008.

\bibitem{jin2012real}
M.~Jin and T.~Maruyama, ``A real-time stereo vision system using a
  tree-structured dynamic programming on fpga,'' in {\em Proceedings of the
  ACM/SIGDA international symposium on Field Programmable Gate Arrays},
  pp.~21--24, 2012.

\bibitem{kamasaka2018fpga}
R.~Kamasaka, Y.~Shibata, and K.~Oguri, ``An fpga-oriented graph cut algorithm
  for accelerating stereo vision,'' in {\em 2018 International Conference on
  ReConFigurable Computing and FPGAs (ReConFig)}, pp.~1--6, IEEE, 2018.

\bibitem{banz2010real}
C.~Banz, S.~Hesselbarth, H.~Flatt, H.~Blume, and P.~Pirsch, ``Real-time stereo
  vision system using semi-global matching disparity estimation: Architecture
  and fpga-implementation,'' in {\em 2010 International Conference on Embedded
  Computer Systems: Architectures, Modeling and Simulation}, pp.~93--101, IEEE,
  2010.

\bibitem{wang2015real}
W.~Wang, J.~Yan, N.~Xu, Y.~Wang, and F.-H. Hsu, ``Real-time high-quality stereo
  vision system in fpga,'' {\em IEEE Transactions on Circuits and Systems for
  Video Technology}, vol.~25, no.~10, pp.~1696--1708, 2015.

\bibitem{cambuim2017hardware}
L.~F. Cambuim, J.~P. Barbosa, and E.~N. Barros, ``Hardware module for
  low-resource and real-time stereo vision engine using semi-global matching
  approach,'' in {\em Proceedings of the 30th Symposium on Integrated Circuits
  and Systems Design: Chip on the Sands}, pp.~53--58, 2017.

\bibitem{rahnama2018r3sgm}
O.~Rahnama, T.~Cavalleri, S.~Golodetz, S.~Walker, and P.~Torr, ``R3sgm:
  Real-time raster-respecting semi-global matching for power-constrained
  systems,'' in {\em 2018 International Conference on Field-Programmable
  Technology (FPT)}, pp.~102--109, IEEE, 2018.

\bibitem{cambuim2019fpga}
L.~F. Cambuim, L.~A. Oliveira, E.~N. Barros, and A.~P. Ferreira, ``An
  fpga-based real-time occlusion robust stereo vision system using semi-global
  matching,'' {\em Journal of Real-Time Image Processing}, pp.~1--22, 2019.

\bibitem{zhao2020fp}
J.~Zhao, T.~Liang, L.~Feng, W.~Ding, S.~Sinha, W.~Zhang, and S.~Shen,
  ``Fp-stereo: Hardware-efficient stereo vision for embedded applications,''
  {\em arXiv preprint arXiv:2006.03250}, 2020.

\bibitem{rahnama2018real}
O.~Rahnama, D.~Frost, O.~Miksik, and P.~H. Torr, ``Real-time dense stereo
  matching with elas on fpga-accelerated embedded devices,'' {\em IEEE Robotics
  and Automation Letters}, vol.~3, no.~3, pp.~2008--2015, 2018.

\bibitem{rahnama2019real}
O.~Rahnama, T.~Cavallari, S.~Golodetz, A.~Tonioni, T.~Joy, L.~Di~Stefano,
  S.~Walker, and P.~H. Torr, ``Real-time highly accurate dense depth on a power
  budget using an fpga-cpu hybrid soc,'' {\em IEEE Transactions on Circuits and
  Systems II: Express Briefs}, vol.~66, no.~5, pp.~773--777, 2019.

\bibitem{hirschmuller2005accurate}
H.~Hirschmuller, ``Accurate and efficient stereo processing by semi-global
  matching and mutual information,'' in {\em 2005 IEEE Computer Society
  Conference on Computer Vision and Pattern Recognition (CVPR'05)}, vol.~2,
  pp.~807--814, IEEE, 2005.

\bibitem{honegger2014real}
D.~Honegger, H.~Oleynikova, and M.~Pollefeys, ``Real-time and low latency
  embedded computer vision hardware based on a combination of fpga and mobile
  cpu,'' in {\em 2014 IEEE/RSJ International Conference on Intelligent Robots
  and Systems}, pp.~4930--4935, IEEE, 2014.

\bibitem{mattoccia2015passive}
S.~Mattoccia and M.~Poggi, ``A passive rgbd sensor for accurate and real-time
  depth sensing self-contained into an fpga,'' in {\em Proceedings of the 9th
  International Conference on Distributed Smart Cameras}, pp.~146--151, 2015.

\bibitem{gehrig2009real}
S.~K. Gehrig, F.~Eberli, and T.~Meyer, ``A real-time low-power stereo vision
  engine using semi-global matching,'' in {\em International Conference on
  Computer Vision Systems}, pp.~134--143, Springer, 2009.

\bibitem{hernandez2016embedded}
D.~Hernandez-Juarez, A.~Chac{\'o}n, A.~Espinosa, D.~V{\'a}zquez, J.~C. Moure,
  and A.~M. L{\'o}pez, ``Embedded real-time stereo estimation via semi-global
  matching on the gpu,'' {\em Procedia Computer Science}, vol.~80,
  pp.~143--153, 2016.

\bibitem{hirschmuller2007evaluation}
H.~Hirschmuller and D.~Scharstein, ``Evaluation of cost functions for stereo
  matching,'' in {\em 2007 IEEE Conference on Computer Vision and Pattern
  Recognition}, pp.~1--8, IEEE, 2007.

\bibitem{shan2012fpga}
Y.~Shan, Z.~Wang, W.~Wang, Y.~Hao, Y.~Wang, K.~Tsoi, W.~Luk, and H.~Yang,
  ``Fpga based memory efficient high resolution stereo vision system for video
  tolling,'' in {\em 2012 International Conference on Field-Programmable
  Technology}, pp.~29--32, IEEE, 2012.

\bibitem{shan2014hardware}
Y.~Shan, Y.~Hao, W.~Wang, Y.~Wang, X.~Chen, H.~Yang, and W.~Luk, ``Hardware
  acceleration for an accurate stereo vision system using mini-census adaptive
  support region,'' {\em ACM Transactions on Embedded Computing Systems
  (TECS)}, vol.~13, no.~4s, pp.~1--24, 2014.

\bibitem{geiger2010efficient}
A.~Geiger, M.~Roser, and R.~Urtasun, ``Efficient large-scale stereo matching,''
  in {\em Asian conference on computer vision}, pp.~25--38, Springer, 2010.

\bibitem{zagoruyko2015learning}
S.~Zagoruyko and N.~Komodakis, ``Learning to compare image patches via
  convolutional neural networks,'' in {\em Proceedings of the IEEE conference
  on computer vision and pattern recognition}, pp.~4353--4361, 2015.

\bibitem{vzbontar2016stereo}
J.~{\v{Z}}bontar and Y.~LeCun, ``Stereo matching by training a convolutional
  neural network to compare image patches,'' {\em The journal of machine
  learning research}, vol.~17, no.~1, pp.~2287--2318, 2016.

\bibitem{luo2016efficient}
W.~Luo, A.~G. Schwing, and R.~Urtasun, ``Efficient deep learning for stereo
  matching,'' in {\em Proceedings of the IEEE Conference on Computer Vision and
  Pattern Recognition}, pp.~5695--5703, 2016.

\bibitem{seki2017sgm}
A.~Seki and M.~Pollefeys, ``Sgm-nets: Semi-global matching with neural
  networks,'' in {\em Proceedings of the IEEE Conference on Computer Vision and
  Pattern Recognition}, pp.~231--240, 2017.

\bibitem{mayer2016large}
N.~Mayer, E.~Ilg, P.~Hausser, P.~Fischer, D.~Cremers, A.~Dosovitskiy, and
  T.~Brox, ``A large dataset to train convolutional networks for disparity,
  optical flow, and scene flow estimation,'' in {\em Proceedings of the IEEE
  conference on computer vision and pattern recognition}, pp.~4040--4048, 2016.

\bibitem{kuzmin2017end}
A.~Kuzmin, D.~Mikushin, and V.~Lempitsky, ``End-to-end learning of cost-volume
  aggregation for real-time dense stereo,'' in {\em 2017 IEEE 27th
  International Workshop on Machine Learning for Signal Processing (MLSP)},
  pp.~1--6, IEEE, 2017.

\bibitem{li2016high}
H.~Li, X.~Fan, L.~Jiao, W.~Cao, X.~Zhou, and L.~Wang, ``A high performance
  fpga-based accelerator for large-scale convolutional neural networks,'' in
  {\em 2016 26th International Conference on Field Programmable Logic and
  Applications (FPL)}, pp.~1--9, IEEE, 2016.

\bibitem{qiu2016going}
J.~Qiu, J.~Wang, S.~Yao, K.~Guo, B.~Li, E.~Zhou, J.~Yu, T.~Tang, N.~Xu,
  S.~Song, {\em et~al.}, ``Going deeper with embedded fpga platform for
  convolutional neural network,'' in {\em Proceedings of the 2016 ACM/SIGDA
  International Symposium on Field-Programmable Gate Arrays}, pp.~26--35, 2016.

\bibitem{guo2017angel}
K.~Guo, L.~Sui, J.~Qiu, J.~Yu, J.~Wang, S.~Yao, S.~Han, Y.~Wang, and H.~Yang,
  ``Angel-eye: A complete design flow for mapping cnn onto embedded fpga,''
  {\em IEEE Transactions on Computer-Aided Design of Integrated Circuits and
  Systems}, vol.~37, no.~1, pp.~35--47, 2017.

\bibitem{yu2018instruction}
J.~Yu, G.~Ge, Y.~Hu, X.~Ning, J.~Qiu, K.~Guo, Y.~Wang, and H.~Yang,
  ``Instruction driven cross-layer cnn accelerator for fast detection on
  fpga,'' {\em ACM Transactions on Reconfigurable Technology and Systems
  (TRETS)}, vol.~11, no.~3, pp.~1--23, 2018.

\bibitem{nakahara2018lightweight}
H.~Nakahara, H.~Yonekawa, T.~Fujii, and S.~Sato, ``A lightweight yolov2: A
  binarized cnn with a parallel support vector regression for an fpga,'' in
  {\em Proceedings of the 2018 ACM/SIGDA International Symposium on
  field-programmable gate arrays}, pp.~31--40, 2018.

\bibitem{belshaw2008high}
M.~S. Belshaw, {\em A high-speed Iterative Closest Point tracker on an FPGA
  platform}.
\newblock PhD thesis, 2008.

\bibitem{williams2017evaluation}
B.~Williams, ``Evaluation of a soc for real-time 3d slam,'' 2017.

\bibitem{van2019fpga}
B.~Van~Hoorick, ``Fpga-based simultaneous localization and mapping (slam) using
  high-level synthesis,'' 2019.

\bibitem{gautier2014real}
Q.~Gautier, A.~Shearer, J.~Matai, D.~Richmond, P.~Meng, and R.~Kastner,
  ``Real-time 3d reconstruction for fpgas: A case study for evaluating the
  performance, area, and programmability trade-offs of the altera opencl sdk,''
  in {\em 2014 International Conference on Field-Programmable Technology
  (FPT)}, pp.~326--329, IEEE, 2014.

\bibitem{bailey2006consistency}
T.~Bailey, J.~Nieto, J.~Guivant, M.~Stevens, and E.~Nebot, ``Consistency of the
  ekf-slam algorithm,'' in {\em 2006 IEEE/RSJ International Conference on
  Intelligent Robots and Systems}, pp.~3562--3568, IEEE, 2006.

\bibitem{mur2015orb}
R.~Mur-Artal, J.~M.~M. Montiel, and J.~D. Tardos, ``Orb-slam: a versatile and
  accurate monocular slam system,'' {\em IEEE transactions on robotics},
  vol.~31, no.~5, pp.~1147--1163, 2015.

\bibitem{montemerlo2002fastslam}
M.~Montemerlo, S.~Thrun, D.~Koller, B.~Wegbreit, {\em et~al.}, ``Fastslam: A
  factored solution to the simultaneous localization and mapping problem,''
  {\em Aaai/iaai}, vol.~593598, 2002.

\bibitem{gu2015fpga}
M.~Gu, K.~Guo, W.~Wang, Y.~Wang, and H.~Yang, ``An fpga-based real-time
  simultaneous localization and mapping system,'' in {\em 2015 International
  Conference on Field Programmable Technology (FPT)}, pp.~200--203, IEEE, 2015.

\bibitem{cadena2016past}
C.~Cadena, L.~Carlone, H.~Carrillo, Y.~Latif, D.~Scaramuzza, J.~Neira, I.~Reid,
  and J.~J. Leonard, ``Past, present, and future of simultaneous localization
  and mapping: Toward the robust-perception age,'' {\em IEEE Transactions on
  robotics}, vol.~32, no.~6, pp.~1309--1332, 2016.

\bibitem{engel2013semi}
J.~Engel, J.~Sturm, and D.~Cremers, ``Semi-dense visual odometry for a
  monocular camera,'' in {\em Proceedings of the IEEE international conference
  on computer vision}, pp.~1449--1456, 2013.

\bibitem{geiger2013vision}
A.~Geiger, P.~Lenz, C.~Stiller, and R.~Urtasun, ``Vision meets robotics: The
  kitti dataset,'' {\em The International Journal of Robotics Research},
  vol.~32, no.~11, pp.~1231--1237, 2013.

\bibitem{burri2016euroc}
M.~Burri, J.~Nikolic, P.~Gohl, T.~Schneider, J.~Rehder, S.~Omari, M.~W.
  Achtelik, and R.~Siegwart, ``The euroc micro aerial vehicle datasets,'' {\em
  The International Journal of Robotics Research}, vol.~35, no.~10,
  pp.~1157--1163, 2016.

\bibitem{newcombe2011kinectfusion}
R.~A. Newcombe, S.~Izadi, O.~Hilliges, D.~Molyneaux, D.~Kim, A.~J. Davison,
  P.~Kohi, J.~Shotton, S.~Hodges, and A.~Fitzgibbon, ``Kinectfusion: Real-time
  dense surface mapping and tracking,'' in {\em 2011 10th IEEE International
  Symposium on Mixed and Augmented Reality}, pp.~127--136, IEEE, 2011.

\bibitem{bonato2009floating}
V.~Bonato, E.~Marques, and G.~A. Constantinides, ``A floating-point extended
  kalman filter implementation for autonomous mobile robots,'' {\em Journal of
  Signal Processing Systems}, vol.~56, no.~1, pp.~41--50, 2009.

\bibitem{tertei2014fpga}
D.~T. Tertei, J.~Piat, and M.~Devy, ``Fpga design and implementation of a
  matrix multiplier based accelerator for 3d ekf slam,'' in {\em 2014
  International Conference on ReConFigurable Computing and FPGAs (ReConFig14)},
  pp.~1--6, IEEE, 2014.

\bibitem{tertei2016fpga}
D.~T. Tertei, J.~Piat, and M.~Devy, ``Fpga design of ekf block accelerator for
  3d visual slam,'' {\em Computers \& Electrical Engineering}, vol.~55,
  pp.~123--137, 2016.

\bibitem{vincke2012real}
B.~Vincke, A.~Elouardi, and A.~Lambert, ``Real time simultaneous localization
  and mapping: towards low-cost multiprocessor embedded systems,'' {\em EURASIP
  Journal on Embedded Systems}, vol.~2012, no.~1, p.~5, 2012.

\bibitem{vincke2014simd}
B.~Vincke, A.~Elouardi, A.~Lambert, and A.~Dine, ``Simd and openmp optimization
  of ekf-slam,'' in {\em 2014 International Conference on Multimedia Computing
  and Systems (ICMCS)}, pp.~712--716, IEEE, 2014.

\bibitem{fang2017fpga}
W.~Fang, Y.~Zhang, B.~Yu, and S.~Liu, ``Fpga-based orb feature extraction for
  real-time visual slam,'' in {\em 2017 International Conference on Field
  Programmable Technology (ICFPT)}, pp.~275--278, IEEE, 2017.

\bibitem{biadgie2014feature}
Y.~Biadgie and K.-A. Sohn, ``Feature detector using adaptive accelerated
  segment test,'' in {\em 2014 International Conference on Information Science
  \& Applications (ICISA)}, pp.~1--4, IEEE, 2014.

\bibitem{calonder2010brief}
M.~Calonder, V.~Lepetit, C.~Strecha, and P.~Fua, ``Brief: Binary robust
  independent elementary features,'' in {\em European conference on computer
  vision}, pp.~778--792, Springer, 2010.

\bibitem{liu2019eslam}
R.~Liu, J.~Yang, Y.~Chen, and W.~Zhao, ``eslam: An energy-efficient accelerator
  for real-time orb-slam on fpga platform,'' in {\em Proceedings of the 56th
  Annual Design Automation Conference 2019}, pp.~1--6, 2019.

\bibitem{schulz2016harris}
V.~H. Schulz, F.~G. Bombardelli, and E.~Todt, ``A harris corner detector
  implementation in soc-fpga for visual slam,'' in {\em Robotics}, pp.~57--71,
  Springer, 2016.

\bibitem{abouzahir2016large}
M.~Abouzahir, A.~Elouardi, S.~Bouaziz, R.~Latif, and A.~Tajer, ``Large-scale
  monocular fastslam2. 0 acceleration on an embedded heterogeneous
  architecture,'' {\em EURASIP Journal on Advances in Signal Processing},
  vol.~2016, no.~1, p.~88, 2016.

\bibitem{abouzahir2018embedding}
M.~Abouzahir, A.~Elouardi, R.~Latif, S.~Bouaziz, and A.~Tajer, ``Embedding slam
  algorithms: Has it come of age?,'' {\em Robotics and Autonomous Systems},
  vol.~100, pp.~14--26, 2018.

\bibitem{boikos2016semi}
K.~Boikos and C.-S. Bouganis, ``Semi-dense slam on an fpga soc,'' in {\em 2016
  26th International Conference on Field Programmable Logic and Applications
  (FPL)}, pp.~1--4, IEEE, 2016.

\bibitem{boikos2017high}
K.~Boikos and C.-S. Bouganis, ``A high-performance system-on-chip architecture
  for direct tracking for slam,'' in {\em 2017 27th International Conference on
  Field Programmable Logic and Applications (FPL)}, pp.~1--7, IEEE, 2017.

\bibitem{boikos2019scalable}
K.~Boikos and C.-S. Bouganis, ``A scalable fpga-based architecture for depth
  estimation in slam,'' in {\em International Symposium on Applied
  Reconfigurable Computing}, pp.~181--196, Springer, 2019.

\bibitem{detone2018superpoint}
D.~DeTone, T.~Malisiewicz, and A.~Rabinovich, ``Superpoint: Self-supervised
  interest point detection and description,'' in {\em Proceedings of the IEEE
  Conference on Computer Vision and Pattern Recognition Workshops},
  pp.~224--236, 2018.

\bibitem{simo2015discriminative}
E.~Simo-Serra, E.~Trulls, L.~Ferraz, I.~Kokkinos, P.~Fua, and F.~Moreno-Noguer,
  ``Discriminative learning of deep convolutional feature point descriptors,''
  in {\em Proceedings of the IEEE International Conference on Computer Vision},
  pp.~118--126, 2015.

\bibitem{radenovic2018fine}
F.~Radenovi{\'c}, G.~Tolias, and O.~Chum, ``Fine-tuning cnn image retrieval
  with no human annotation,'' {\em IEEE transactions on pattern analysis and
  machine intelligence}, vol.~41, no.~7, pp.~1655--1668, 2018.

\bibitem{knuthwebsite}
Xilinx, ``Dpu for convolutional neural network.''

\bibitem{xu2020cnn}
Z.~Xu, J.~Yu, C.~Yu, H.~Shen, Y.~Wang, and H.~Yang, ``Cnn-based feature-point
  extraction for real-time visual slam on embedded fpga,'' in {\em 2020 IEEE
  28th Annual International Symposium on Field-Programmable Custom Computing
  Machines (FCCM)}, pp.~33--37, IEEE, 2020.

\bibitem{han2015deep}
S.~Han, H.~Mao, and W.~J. Dally, ``Deep compression: Compressing deep neural
  networks with pruning, trained quantization and huffman coding,'' {\em arXiv
  preprint arXiv:1510.00149}, 2015.

\bibitem{krishnan2019quantized}
S.~Krishnan, S.~Chitlangia, M.~Lam, Z.~Wan, A.~Faust, and V.~J. Reddi,
  ``Quantized reinforcement learning (quarl),'' {\em arXiv preprint
  arXiv:1910.01055}, 2019.

\bibitem{langroudi2020adaptive}
H.~F. Langroudi, V.~Karia, J.~L. Gustafson, and D.~Kudithipudi, ``Adaptive
  posit: Parameter aware numerical format for deep learning inference on the
  edge,'' in {\em Proceedings of the IEEE/CVF Conference on Computer Vision and
  Pattern Recognition Workshops}, pp.~726--727, 2020.

\bibitem{tambe2020algorithm}
T.~Tambe, E.-Y. Yang, Z.~Wan, Y.~Deng, V.~J. Reddi, A.~Rush, D.~Brooks, and
  G.-Y. Wei, ``Algorithm-hardware co-design of adaptive floating-point
  encodings for resilient deep learning inference,'' in {\em 2020 57th ACM/IEEE
  Design Automation Conference (DAC)}, pp.~1--6, IEEE, 2020.

\bibitem{li2016ternary}
F.~Li, B.~Zhang, and B.~Liu, ``Ternary weight networks,'' {\em arXiv preprint
  arXiv:1605.04711}, 2016.

\bibitem{choi2019accurate}
J.~Choi, S.~Venkataramani, V.~Srinivasan, K.~Gopalakrishnan, Z.~Wang, and
  P.~Chuang, ``Accurate and efficient 2-bit quantized neural networks,'' in
  {\em Proceedings of the 2nd SysML Conference}, vol.~2019, 2019.

\bibitem{kim2020position}
J.~Kim, K.~Yoo, and N.~Kwak, ``Position-based scaled gradient for model
  quantization and pruning,'' {\em Advances in Neural Information Processing
  Systems}, vol.~33, 2020.

\bibitem{tambe2019adaptivfloat}
T.~Tambe, E.-Y. Yang, Z.~Wan, Y.~Deng, V.~J. Reddi, A.~Rush, D.~Brooks, and
  G.-Y. Wei, ``Adaptivfloat: A floating-point based data type for resilient
  deep learning inference,'' {\em arXiv preprint arXiv:1909.13271}, 2019.

\bibitem{yu2020cnn}
J.~Yu, F.~Gao, J.~Cao, C.~Yu, Z.~Zhang, Z.~Huang, Y.~Wang, and H.~Yang,
  ``Cnn-based monocular decentralized slam on embedded fpga,'' 2020.

\bibitem{zhan2018unsupervised}
H.~Zhan, R.~Garg, C.~Saroj~Weerasekera, K.~Li, H.~Agarwal, and I.~Reid,
  ``Unsupervised learning of monocular depth estimation and visual odometry
  with deep feature reconstruction,'' in {\em Proceedings of the IEEE
  Conference on Computer Vision and Pattern Recognition}, pp.~340--349, 2018.

\bibitem{arandjelovic2016netvlad}
R.~Arandjelovic, P.~Gronat, A.~Torii, T.~Pajdla, and J.~Sivic, ``Netvlad: Cnn
  architecture for weakly supervised place recognition,'' in {\em Proceedings
  of the IEEE conference on computer vision and pattern recognition},
  pp.~5297--5307, 2016.

\bibitem{yuinca}
J.~Yu, Z.~Xu, S.~Zeng, C.~Yu, J.~Qiu, C.~Shen, Y.~Xu, G.~Dai, Y.~Wang, and
  H.~Yang, ``Inca: Interruptible cnn accelerator for multi-tasking in embedded
  robots,'' in {\em 2020 57th ACM/ESDA/IEEE Design Automation Conference
  (DAC)}, IEEE, 2020.

\bibitem{mur2017orb}
R.~Mur-Artal and J.~D. Tard{\'o}s, ``Orb-slam2: An open-source slam system for
  monocular, stereo, and rgb-d cameras,'' {\em IEEE Transactions on Robotics},
  vol.~33, no.~5, pp.~1255--1262, 2017.

\bibitem{liu2020eavr}
S.~Liu, {\em Engineering Autonomous Vehicles and Robots: The DragonFly
  Modular‐based Approach}.
\newblock Wiley-IEEE Press, 1~ed., 3 2020.

\bibitem{maimone2007two}
M.~Maimone, Y.~Cheng, and L.~Matthies, ``Two years of visual odometry on the
  mars exploration rovers,'' {\em Journal of Field Robotics}, vol.~24, no.~3,
  pp.~169--186, 2007.

\bibitem{klingner2013street}
B.~Klingner, D.~Martin, and J.~Roseborough, ``Street view
  motion-from-structure-from-motion,'' in {\em Proceedings of the IEEE
  International Conference on Computer Vision}, pp.~953--960, 2013.

\bibitem{jeong2011pushing}
Y.~Jeong, D.~Nister, D.~Steedly, R.~Szeliski, and I.-S. Kweon, ``Pushing the
  envelope of modern methods for bundle adjustment,'' {\em IEEE transactions on
  pattern analysis and machine intelligence}, vol.~34, no.~8, pp.~1605--1617,
  2011.

\bibitem{wu2011multicore}
C.~Wu, S.~Agarwal, B.~Curless, and S.~M. Seitz, ``Multicore bundle
  adjustment,'' in {\em CVPR 2011}, pp.~3057--3064, IEEE, 2011.

\bibitem{eriksson2016consensus}
A.~Eriksson, J.~Bastian, T.-J. Chin, and M.~Isaksson, ``A consensus-based
  framework for distributed bundle adjustment,'' in {\em Proceedings of the
  IEEE Conference on Computer Vision and Pattern Recognition}, pp.~1754--1762,
  2016.

\bibitem{zhang2017distributed}
R.~Zhang, S.~Zhu, T.~Fang, and L.~Quan, ``Distributed very large scale bundle
  adjustment by global camera consensus,'' in {\em Proceedings of the IEEE
  International Conference on Computer Vision}, pp.~29--38, 2017.

\bibitem{suleiman2019navion}
A.~Suleiman, Z.~Zhang, L.~Carlone, S.~Karaman, and V.~Sze, ``Navion: A 2-mw
  fully integrated real-time visual-inertial odometry accelerator for
  autonomous navigation of nano drones,'' {\em IEEE Journal of Solid-State
  Circuits}, vol.~54, no.~4, pp.~1106--1119, 2019.

\bibitem{liu2020pi}
Q.~Liu, S.~Qin, B.~Yu, J.~Tang, and S.~Liu, ``$\pi$-ba: Bundle adjustment
  hardware accelerator based on distribution of 3d-point observations,'' {\em
  IEEE Transactions on Computers}, 2020.

\bibitem{sun2020bax}
R.~Sun, P.~Liu, J.~Xue, S.~Yang, J.~Qian, and R.~Ying, ``Bax: A bundle
  adjustment accelerator with decoupled access/execute architecture for visual
  odometry,'' {\em IEEE Access}, vol.~8, pp.~75530--75542, 2020.

\bibitem{leven2002framework}
P.~Leven and S.~Hutchinson, ``A framework for real-time path planning in
  changing environments,'' {\em The International Journal of Robotics
  Research}, vol.~21, no.~12, pp.~999--1030, 2002.

\bibitem{karaman2011sampling}
S.~Karaman and E.~Frazzoli, ``Sampling-based algorithms for optimal motion
  planning,'' {\em The international journal of robotics research}, vol.~30,
  no.~7, pp.~846--894, 2011.

\bibitem{gammell2015batch}
J.~D. Gammell, S.~S. Srinivasa, and T.~D. Barfoot, ``Batch informed trees
  (bit*): Sampling-based optimal planning via the heuristically guided search
  of implicit random geometric graphs,'' in {\em 2015 IEEE international
  conference on robotics and automation (ICRA)}, pp.~3067--3074, IEEE, 2015.

\bibitem{hauser2015lazy}
K.~Hauser, ``Lazy collision checking in asymptotically-optimal motion
  planning,'' in {\em 2015 IEEE International Conference on Robotics and
  Automation (ICRA)}, pp.~2951--2957, IEEE, 2015.

\bibitem{yershova2007improving}
A.~Yershova and S.~M. LaValle, ``Improving motion-planning algorithms by
  efficient nearest-neighbor searching,'' {\em IEEE Transactions on Robotics},
  vol.~23, no.~1, pp.~151--157, 2007.

\bibitem{wang2015fast}
W.~Wang, D.~Balkcom, and A.~Chakrabarti, ``A fast online spanner for roadmap
  construction,'' {\em The International Journal of Robotics Research},
  vol.~34, no.~11, pp.~1418--1432, 2015.

\bibitem{murray2019programmable}
S.~Murray, W.~Floyd-Jones, G.~Konidaris, and D.~J. Sorin, ``A programmable
  architecture for robot motion planning acceleration,'' in {\em 2019 IEEE 30th
  International Conference on Application-specific Systems, Architectures and
  Processors (ASAP)}, vol.~2160, pp.~185--188, IEEE, 2019.

\bibitem{bialkowski2011massively}
J.~Bialkowski, S.~Karaman, and E.~Frazzoli, ``Massively parallelizing the rrt
  and the rrt,'' in {\em 2011 IEEE/RSJ International Conference on Intelligent
  Robots and Systems}, pp.~3513--3518, IEEE, 2011.

\bibitem{pan2012gpu}
J.~Pan and D.~Manocha, ``Gpu-based parallel collision detection for fast motion
  planning,'' {\em The International Journal of Robotics Research}, vol.~31,
  no.~2, pp.~187--200, 2012.

\bibitem{pan2010g}
J.~Pan, C.~Lauterbach, and D.~Manocha, ``g-planner: Real-time motion planning
  and global navigation using gpus.,'' in {\em AAAI}, 2010.

\bibitem{atay2006motion}
N.~Atay and B.~Bayazit, ``A motion planning processor on reconfigurable
  hardware,'' in {\em Proceedings 2006 IEEE International Conference on
  Robotics and Automation, 2006. ICRA 2006.}, pp.~125--132, IEEE, 2006.

\bibitem{murray2016microarchitecture}
S.~Murray, W.~Floyd-Jones, Y.~Qi, G.~Konidaris, and D.~J. Sorin, ``The
  microarchitecture of a real-time robot motion planning accelerator,'' in {\em
  2016 49th Annual IEEE/ACM International Symposium on Microarchitecture
  (MICRO)}, pp.~1--12, IEEE, 2016.

\bibitem{lian2018dadu}
S.~Lian, Y.~Han, X.~Chen, Y.~Wang, and H.~Xiao, ``Dadu-p: A scalable
  accelerator for robot motion planning in a dynamic environment,'' in {\em
  2018 55th ACM/ESDA/IEEE Design Automation Conference (DAC)}, pp.~1--6, IEEE,
  2018.

\bibitem{bondhugula2006hardware}
U.~Bondhugula, A.~Devulapalli, J.~Dinan, J.~Fernando, P.~Wyckoff, E.~Stahlberg,
  and P.~Sadayappan, ``Hardware/software integration for fpga-based all-pairs
  shortest-paths,'' in {\em 2006 14th Annual IEEE Symposium on
  Field-Programmable Custom Computing Machines}, pp.~152--164, IEEE, 2006.

\bibitem{sridharan2009hardware}
K.~Sridharan, T.~Priya, and P.~R. Kumar, ``Hardware architecture for finding
  shortest paths,'' in {\em TENCON 2009-2009 IEEE Region 10 Conference},
  pp.~1--5, IEEE, 2009.

\bibitem{takei2015evaluation}
Y.~Takei, M.~Hariyama, and M.~Kameyama, ``Evaluation of an fpga-based
  shortest-path-search accelerator,'' in {\em Proceedings of the International
  Conference on Parallel and Distributed Processing Techniques and Applications
  (PDPTA)}, p.~613, The Steering Committee of The World Congress in Computer
  Science, Computer Engineering and Applied Computing (WorldComp), 2015.

\bibitem{vipin2018fpga}
K.~Vipin and S.~A. Fahmy, ``Fpga dynamic and partial reconfiguration: A survey
  of architectures, methods, and applications,'' {\em ACM Computing Surveys
  (CSUR)}, vol.~51, no.~4, pp.~1--39, 2018.

\bibitem{liu2010minimizing}
S.~Liu, R.~N. Pittman, and A.~Forin, ``Minimizing partial reconfiguration
  overhead with fully streaming dma engines and intelligent icap controller,''
  in {\em FPGA}, p.~292, Citeseer, 2010.

\bibitem{liu2013achieving}
S.~Liu, R.~N. Pittman, A.~Forin, and J.-L. Gaudiot, ``Achieving energy
  efficiency through runtime partial reconfiguration on reconfigurable
  systems,'' {\em ACM Transactions on Embedded Computing Systems (TECS)},
  vol.~12, no.~3, p.~72, 2013.

\bibitem{rublee2011orb}
E.~Rublee, V.~Rabaud, K.~Konolige, and G.~R. Bradski, ``Orb: An efficient
  alternative to sift or surf.,'' in {\em ICCV}, vol.~11, p.~2, Citeseer, 2011.

\bibitem{lucas1981iterative}
B.~D. Lucas and T.~Kanade, ``{An iterative image registration technique with an
  application to stereo vision},'' in {\em Proceedings of the 7th International
  Joint Conference on Artificial Intelligence}, 1981.

\bibitem{fang2018dragonfly+}
W.~Fang, Y.~Zhang, B.~Yu, and S.~Liu, ``Dragonfly+: Fpga-based quad-camera
  visual slam system for autonomous vehicles,'' {\em Proc. IEEE HotChips},
  p.~1, 2018.

\bibitem{qin2018vins}
T.~Qin, P.~Li, and S.~Shen, ``Vins-mono: A robust and versatile monocular
  visual-inertial state estimator,'' {\em IEEE Transactions on Robotics},
  vol.~34, no.~4, pp.~1004--1020, 2018.

\bibitem{sun2018msckf}
K.~{Sun}, K.~{Mohta}, B.~{Pfrommer}, M.~{Watterson}, S.~{Liu}, Y.~{Mulgaonkar},
  C.~J. {Taylor}, and V.~{Kumar}, ``Robust stereo visual inertial odometry for
  fast autonomous flight,'' {\em IEEE Robotics and Automation Letters}, vol.~3,
  no.~2, pp.~965--972, 2018.

\bibitem{szeliski2010computer}
R.~Szeliski, {\em Computer Vision: Algorithms and Applications}.
\newblock Texts in Computer Science, Springer London, 2010.

\bibitem{Elas2010Geiger}
A.~Geiger, M.~Roser, and R.~Urtasun, ``Efficient large-scale stereo matching,''
  in {\em Proceedings of the 10th Asian Conference on Computer Vision}, 2010.

\bibitem{feng2019stereo}
Y.~Feng, P.~Whatmough, and Y.~Zhu, ``Asv: Accelerated stereo vision system,''
  in {\em Proceedings of the 52nd Annual IEEE/ACM International Symposium on
  Microarchitecture}, MICRO '52, p.~643–656, 2019.

\bibitem{Hen2015kcf}
J.~F. Henriques, R.~Caseiro, P.~Martins, and J.~Batista, ``High-speed tracking
  with kernelized correlation filters,'' {\em IEEE Transactions on Pattern
  Analysis \& Machine Intelligence}, vol.~37, pp.~583--596, mar 2015.

\bibitem{kelly2013mobile}
A.~Kelly, {\em Mobile Robotics: Mathematics, Models, and Methods}.
\newblock Cambridge University Press, 2013.

\bibitem{tang2018pisoc}
J.~{Tang}, B.~{Yu}, S.~{Liu}, Z.~{Zhang}, W.~{Fang}, and Y.~{Zhang},
  ``{\(\pi\)}-soc: Heterogeneous soc architecture for visual inertial slam
  applications,'' in {\em 2018 IEEE/RSJ International Conference on Intelligent
  Robots and Systems (IROS)}, pp.~8302--8307, 2018.

\bibitem{qin2019ba}
S.~Qin, Q.~Liu, B.~Yu, and S.~Liu, ``{\(\pi\)}-ba: Bundle adjustment
  acceleration on embedded fpgas with co-observation optimization,'' in {\em
  27th {IEEE} Annual International Symposium on Field-Programmable Custom
  Computing Machines, {FCCM} 2019, San Diego, CA, USA, April 28 - May 1, 2019},
  pp.~100--108, {IEEE}, 2019.

\bibitem{baldataset}
\url{https://grail.cs.washington.edu/projects/bal/}.

\bibitem{mckerracher1992design}
P.~L. Mckerracher, R.~P. Cain, J.~C. Barnett, W.~S. Green, and J.~D. Kinnison,
  ``Design and test of field programmable gate arrays in space applications,''
  1992.

\bibitem{berg2019fpga}
M.~Berg, ``Fpga mitigation strategies for critical applications,'' 2019.

\bibitem{sheldonflash}
D.~Sheldon, ``Flash-based fpga nepp fy12 summary report,''

\bibitem{gaillard2011single}
R.~Gaillard, ``Single event effects: Mechanisms and classification,'' in {\em
  Soft errors in modern electronic systems}, pp.~27--54, Springer, 2011.

\bibitem{wirthlin2013fpgas}
M.~Wirthlin, ``Fpgas operating in a radiation environment: lessons learned from
  fpgas in space,'' {\em Journal of Instrumentation}, vol.~8, no.~02,
  p.~C02020, 2013.

\bibitem{brosser2014seu}
F.~Brosser and E.~Milh, ``Seu mitigation techniques for advanced reprogrammable
  fpga in space,'' Master's thesis, 2014.

\bibitem{ahmed2016fault}
B.~Ahmed and C.~Basha, {\em Fault mitigation strategies for reliable FPGA
  architectures}.
\newblock PhD thesis, Rennes 1, 2016.

\bibitem{habinc2002suitability}
S.~Habinc, ``Suitability of reprogrammable fpgas in space applications,'' {\em
  Gaisler Research,” Feasibility Report}, 2002.

\bibitem{lentaris2018high}
G.~Lentaris, K.~Maragos, I.~Stratakos, L.~Papadopoulos, O.~Papanikolaou,
  D.~Soudris, M.~Lourakis, X.~Zabulis, D.~Gonzalez-Arjona, and G.~Furano,
  ``High-performance embedded computing in space: Evaluation of platforms for
  vision-based navigation,'' {\em Journal of Aerospace Information Systems},
  vol.~15, no.~4, pp.~178--192, 2018.

\bibitem{li2019enabling}
T.~Y. Li and S.~Liu, ``Enabling commercial autonomous robotic space
  explorers,'' {\em IEEE Potentials}, vol.~39, no.~1, pp.~29--36, 2019.

\bibitem{ratter2004fpgas}
D.~Ratter, ``Fpgas on mars,'' {\em Xcell J}, vol.~50, pp.~8--11, 2004.

\bibitem{bell2003mars}
J.~F. Bell~III, S.~Squyres, K.~E. Herkenhoff, J.~Maki, H.~Arneson, D.~Brown,
  S.~Collins, A.~Dingizian, S.~Elliot, E.~Hagerott, {\em et~al.}, ``Mars
  exploration rover athena panoramic camera (pancam) investigation,'' {\em
  Journal of Geophysical Research: Planets}, vol.~108, no.~E12, 2003.

\bibitem{sfsdet}
``Space flight system design and environmental test.''
  \url{https://www.nasa.gov/sites/default/files/atoms/files/std8070.1.pdf}.
\newblock Accessed: 2020-09-01.

\bibitem{malin2017mars}
M.~C. Malin, M.~A. Ravine, M.~A. Caplinger, F.~Tony~Ghaemi, J.~A. Schaffner,
  J.~N. Maki, J.~F. Bell~III, J.~F. Cameron, W.~E. Dietrich, K.~S. Edgett, {\em
  et~al.}, ``The mars science laboratory (msl) mast cameras and descent imager:
  investigation and instrument descriptions,'' {\em Earth and Space Science},
  vol.~4, no.~8, pp.~506--539, 2017.

\bibitem{edwards2003electra}
C.~D. Edwards, T.~C. Jedrey, A.~Devereaux, R.~DePaula, and M.~Dapore, ``The
  electra proximity link payload for mars relay telecommunications and
  navigation,'' 2003.

\bibitem{johnson2017lander}
A.~Johnson, S.~Aaron, J.~Chang, Y.~Cheng, J.~Montgomery, S.~Mohan,
  S.~Schroeder, B.~Tweddle, N.~Trawny, and J.~Zheng, ``The lander vision system
  for mars 2020 entry descent and landing,'' 2017.

\bibitem{vivado}
``Vivado high-level synthesis.''
  \url{https://www.xilinx.com/products/design-tools/vivado/integration/esl-design.html}.
\newblock Accessed: 2020-09-10.

\end{thebibliography}

%








\end{document}